\documentclass[letterpaper, 10 pt, conference]{ieeeconf}  

\IEEEoverridecommandlockouts                              

\usepackage{graphics} 
\usepackage{epsfig} 
\usepackage{mathptmx} 
\usepackage{times} 
\usepackage{amsmath} 
\usepackage{amssymb}  
\usepackage{multirow}
\usepackage{bbm}
\usepackage{float}
\usepackage{tabularx}
\usepackage{subcaption}
\usepackage{booktabs}
\usepackage{gensymb}
\usepackage{url}
\usepackage{hyperref}
\hypersetup{
    colorlinks=true,
    linkcolor=blue,
    filecolor=blue,      
    urlcolor=blue,
    pdftitle={SVG-Net: An SVG-based Trajectory Prediction Model},
    pdfpagemode=FullScreen,
    }
\title{\LARGE \bf
SVG-Net: An SVG-based Trajectory Prediction Model
}

\author{Mohammadhossein Bahari$^{1}$, Vahid Zehtab$^{*,2}$, Sadegh Khorasani$^{*,2}$, Sana Ayromlou$^{*,2}$, \\ Saeed Saadatnejad$^{1}$, and Alexandre Alahi$^{1}$ \\
$^{1}$ EPFL, $^2${Sharif University of Technology} \\
        {\tt\small mohammadhossein.bahari@epfl.ch}
}

\begin{document}

\maketitle
\let\thefootnote\relax\footnotetext{\leftline{$^*$ Equal contribution as the second authors. }}

\begin{abstract}
Anticipating motions of vehicles in a scene is an essential problem for safe autonomous driving systems. To this end, the comprehension of the scene's infrastructure is often the main clue for predicting future trajectories.
Most of the proposed approaches represent the scene with a rasterized format and some of the more recent approaches leverage custom vectorized formats.
In contrast, we propose representing the scene's information by employing Scalable Vector Graphics (SVG). 
SVG is a well-established format that matches the problem of trajectory prediction better than rasterized formats while being more general than arbitrary vectorized formats.
SVG has the potential to provide the convenience and generality of raster-based solutions if coupled with a powerful tool such as CNNs, for which we introduce SVG-Net. SVG-Net is a Transformer-based Neural Network that can effectively capture the scene's information from SVG inputs. Thanks to the self-attention mechanism in its Transformers, SVG-Net can also adequately apprehend relations amongst the scene and the agents. We demonstrate SVG-Net's effectiveness by evaluating its performance on the publicly-available Argoverse forecasting dataset. Finally, we illustrate how, by using SVG, one can benefit from datasets and advancements in other research fronts that also utilize the same input format. Our code is available at \href{https://vita-epfl.github.io/SVGNet/}{https://vita-epfl.github.io/SVGNet/}.
 
\end{abstract}

\section{Introduction}
Autonomous driving is not exclusively a perception nor a planning problem. A prediction pillar (in charge of predicting vehicles' trajectories) is essential and has been overlooked for years.
It is only recently that the community is proposing dataset \cite{Argoverse,nuscenes2019,lyft2020} and challenges  \cite{waymo_challenge,argochall_2020} to move the field forward. We start to see an arm race of methods studying popular architecture designs such as Convolutional Neural Networks (CNN) to solve the prediction task \cite{ren2021safety,salzmann2020trajectron,sadeghian2018car} using rasterized inputs. 
In this paper, we show an alternative to the common practices to move the field forward.
We propose to tackle the vehicle trajectory prediction task with a new input representation, Scalable Vector Graphics (SVG), which better suits the problem's characteristics. 
We intentionally emphasize on the use of a \textit{standard} vectorized image format to ease its usage and leverage related research dealing with such a format.

\begin{figure}[t]
    \centering
    \includegraphics[width=\linewidth]{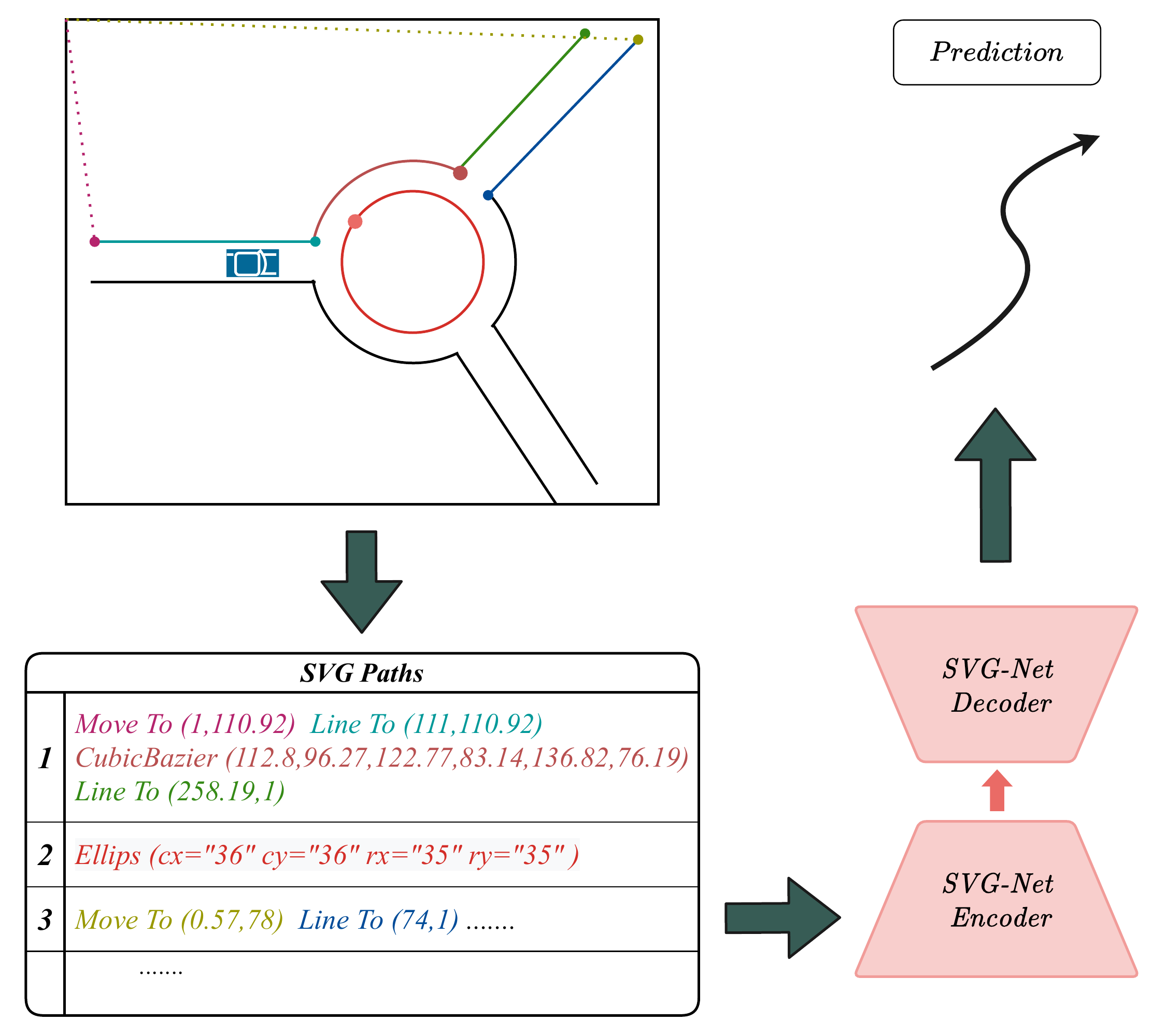}
    \caption{Illustration of how our SVG-Net model works. 
    On the left, the SVG representation of a scene is depicted. SVG consists of a set of paths that each describe a part of the image. The paths are made of different commands with different graphical functionalities. The attributes of the commands are written in parenthesis. SVG-Net, our encoder-decoder network, takes SVG representation as input and predicts future positions of the vehicle.}
    \label{fig:svg_concept}
\end{figure}

Trajectory prediction has been studied widely with neural networks to forecast human social interactions \cite{alahi2016sociallstm,vemula2018socialattention,ivanovic2019trajectron}. 
In human trajectory prediction, the static context is often discarded and the focus is on best modeling the interactions with other agents \cite{saadatnejad2021socially,noticing2021,vemula2018socialattention}. However, for vehicles, the scene context (road infrastructure) has as much (if not more) significance as other agents' dynamic and contributes a major part in the challenges of the task. 

The scene's context information can be represented in either a rasterized format as in \cite{nuscenes2019} or a vectorized representation, similar to \cite{Argoverse}.  By taking the input domain structure of neural networks in mind, it is clear that the scene's contextual information should be represented in a way that is digestible for the networks. 
In most of the previous works on vehicle trajectory prediction \cite{park2020diverse,salzmann2020trajectron,hong2019rules,chai2019multipath,Chauffeurnet_waymo}, scene's contextual information is rendered into image-like raster inputs, and 2D  Convolutional Neural Networks (CNN) are employed to learn an abstract representation. This is inspired by the success of CNNs in various computer vision tasks  \cite{huang2017densely,he2016deep,girshick2015fast,redmon2016you}, making rendered images and CNNs as standard input representations and processors, respectively. 
 
There exist two disadvantages for using rasterized inputs for the trajectory prediction task. First, in vectorized formats, the information is usually structured and the conversion to rasterized inputs would mean losing some of the structural information. Second, while the scene information 
suffices to understand the geometry of the scene, images are of high dimensions and potentially inefficient to learn from. 
To mitigate these issues, researchers recently proposed direct use of vectorized data as inputs \cite{gao2020_waymo_vectornet,liang2020_uber_learning}. However, due to the use of a non-standard data format, these approaches are limited to their own model designs.

In summary, we believe that an ideal representation for the scene information should address the following concerns: 
\begin{enumerate}
    \item it should be able to preserve the available structure in the data in an efficient manner,
    \item it should be such that neural networks can process it,
    \item it should be a standardized representation.
\end{enumerate}
While the importance of the first two points is obvious, the necessity of using a standard representation is to provide ease of use and, more importantly, to enable researchers to accumulate knowledge over different computer vision tasks that employ the same standard representation.

In this work, we show that it is possible to interpret the vectorized information as Scalable Vector Graphics (SVG) \cite{w3c_svg} to represent the scenes' information. SVG inputs satisfy all the three mentioned necessities. First, SVG's inherent format allows densely preserving the structural representation of the vectorized data. Second, it has been shown recently that neural networks are able to process SVG \cite{carlier2020deepsvg}. Finally, it is widely used in digital graphics and web-animations making it a standard format which motivated many previous researches to study this representation \cite{das2021cloud2curve,li2020differentiable,liu2017raster,carlier2020deepsvg,lopes2019learned}.
Hence, we present a transformer-based neural network that learns a latent representation from SVG to solve the prediction task, referred to as SVG-Net.

Figure \ref{fig:svg_concept} displays the overall framework proposed in this work. SVG-represented inputs are taken by SVG-Net with an encoder-decoder architecture to forecast future positions. The scene's information and the states of agents are transmitted to a latent space by the encoders. Our decoder then extracts the relations in the input utilizing a self-attention mechanism. We demonstrate the effectiveness of our approach using the Argoverse dataset \cite{chang2019argoverse} as a well-known yet challenging public dataset. We finally share our source code to help promote our aim for a new standardized paradigm. \\
The contributions of this work are as follows: 
\begin{itemize}
    \item we propose to use the standardized SVG representation rather than raster images in vehicle trajectory prediction,
    \item we propose SVG-Net, a transformer-based network which can effectively perceive SVG-represented scenes and agents' interactions,
    \item we share all the code to reproduce our experiments on a publicly available dataset to move the field forward.
\end{itemize}

\section{Previous works}
\noindent\textbf{Vehicle trajectory prediction.}
Vehicle trajectory prediction has been first addressed by means of knowledge-driven methods. To incorporate scene's information, \cite{benz} proposed associating vehicle's positions with the lanes of the road. The leader-follower model is proposed in \cite{leader-follower} to address the interactions among agents. Authors in \cite{alahi2016social} showed the boost over the well-known model, Social Forces \cite{social-forces} in learning interactions between humans by a social Long-Short Term Memory (LSTM) network. Since then, many works used neural networks to understand the intents of interacting agents. Multiple pooling approaches are used in \cite{shi2021sgcn,salzmann2020trajectron,kothari2020human,bartoli2018context,alahi2016sociallstm} to share features of agents between them in order to have a socially-compliant prediction. 
In this work, we benefit from the self-attention mechanism to learn the relations between the agents and the scene simultaneously. 

Vehicles are constrained to the roads. Therefore, understanding the scene plays an important role in vehicle trajectory prediction.
In \cite{salzmann2020trajectron,park2020diverse,sadeghian2018car}, a Convolutional Neural Network (CNN) is used to extract scene's features.
The learning power of CNN is utilized in  \cite{ren2021safety,hong2019rules,chai2019multipath,Chauffeurnet_waymo} to implicitly learn both interactions and the scene semantics. To this end, they render the scene semantics and states of agents in the scene in a multidimensional image and use CNNs to capture the underlying relations between dimensions. While CNNs showed impressive capabilities in the classification task \cite{xie2020self,huang2017densely,he2016deep}, we argue that map information has sparse and structured information which can be represented in more suitable formats, such as SVG. This would also help the model with better understanding the environment.

\noindent\textbf{Representations for trajectory prediction.}
While raster representation is the most common representation employed by the previous work, few have studied using other standardized representations for trajectory prediction. Authors in \cite{curbside} learn motion primitives in the Curbside coordinate frame. In \cite{frenet}, instead of the Cartesian coordinate, the Fren\'et coordinate frame is employed to represent vehicles' states. 
Recently, \cite{gao2020_waymo_vectornet,liang2020_uber_learning} proposed the use of Graph Neural Networks and for scene semantics to be represented as connected graphs. Their methods require the input to be represented in a vectorized format
which is then converted to their own defined graph representation. In contrast, our input representation is of a well explored standardized format which allows benefiting from advancements in its related research and ease of use.

\noindent\textbf{Vector graphics representation learning.}
Although vector graphics have become popular in different applications, it has not received much attention in the deep learning community over the last decade, in contrast to its counterpart, raster images. In recent years, however, there is a growing research interest in vector graphics. Image vectorization \textit{i.e.,} converting raster images to vectors, has been addressed in \cite{liu2017raster,guo2019deepline} using deep learning techniques. Vectorization of technical line
drawings is studied in \cite{egiazarian2020deepVectorization}. They first remove image artifacts and then leverage a transformer-based network to estimate vector primitives. 
Closer to our work are the studies that learn a latent representation from vector graphics, usually leading to vector graphics generation. In \cite{lopes2019learned} an LSTM-based model is employed to generate vector graphics. 
Authors in \cite{reddy2021im2vec} employed the differentiable rasterizer \cite{li2020differentiable} to generate vector graphics with raster-only supervisions. 
Recently, DeepSVG \cite{carlier2020deepsvg} leveraged a hierarchical structure to generate SVG objects with a Variational Auto-Encoder (VAE) \cite{kingma2013auto}. Their work shows that SVG can be perceived in a hierarchical manner with deep neural networks.  Motivated by DeepSVG, we utilize SVG as the input representation and take advantage of all these recent findings.

\section{Method}
\subsection{Input representation}
\subsubsection{Vector graphics and SVG}
Scalable Vector Graphics (SVG) is a standard vectorized representation of visual content. By providing preliminary tools to form fundamental geometrical shapes and graphics through vectors, SVG can encode images and animations into XML objects \cite{w3c_svg}. This way, it can fully support seamless transformations (\textit{e.g.}, scaling, rotation, ...) that could be problematic for pixel-based content representations by nature. Coupled with its vast capabilities and ease of use, SVG is a suitable representation for most scale-invariant and raster-like graphics.

The simplest SVG objects consist of a hierarchy of permutation invariant path objects. Each path describes a geometric shape (\textit{e.g.}, a closed or open curve)
through XML attributes and a sequence of SVG commands, which further express the shape's underlying geometry. 
Figure \ref{fig:svg_concept} has an illustration on how SVG representation works. It consists of several paths where each path represents part of the image by a set of commands and each command has its specific attributes. Although the Scalable Vector Graphics' API supports various and higher-level graphical descriptions (\textit{e.g.}, circle, rectangle, ...), most SVG objects can be simplified to follow the called format without losing much to any expressivity. Therefore, in this paper, SVG is assumed to be only a collection of SVG path objects.

\subsubsection{Representation of the scene and trajectories}

The inputs for trajectory prediction comprises three components: scene semantics ($S$), an arbitrary number of frames of the main agent's past trajectory, and other agents' preceding trajectories. At any time frame $t$, the $i$-th agent is represented by its xy-coordinates $(x_i^t, y_i^t)$. Hence, the observation sequence for agent $i$ would be $ X_i = \{(x_i^t, y_i^t)\}_{t=1}^{t=T_{obs}},$
where $T_{obs}$ is the number of observation frames. The goal is to predict the next $T_{pred}$ frames of the main agent as
$Y_i = \{(x^t_i, y^t_i)\}_{t=T_{obs} + 1}^{t=T_{obs}+T_{pred}}$.
Without loss of generality, we represent the main agent as the first index and the rest of agents in the scene as other indices.

Most of the recent datasets store scene information using piece-wise linearly approximated vectors. We translated these vectors to the SVG format by using the Line SVG command; to 
represent each vector with a Line command.
After the conversion of scene semantics to the SVG format, to feed the resulting representation of the scene to deep neural networks and support a wide range of SVG objects, we follow the approach suggested by \cite{carlier2020deepsvg}. For each path object $P_i$ with $n_i$ commands, the $k$-th command is represented as $C^k_{P_i}$, a vector with predefined length holding the command's type and its corresponding arguments. Therefore, each path object is defined as a list of its representative command vectors $\{C^k_{P_i}\}_{k=1}^{k=n_i}$. Formally, 
    $Scene \rightarrow \{P_i\}_{i=1}^{i=N_P} = \{\{C^k_{P_i}\}_{k=1}^{k=n_i}\}_{i=1}^{i=N_P},$
describes the SVG representation where $N_P$ is the number of SVG path objects in the scene's SVG. This formulation allows the direct and computationally cheap conversion of samples from datasets that store the vectorized information of the scene to SVG-Net digestible SVG inputs while also providing support for Image Tracing based SVG conversions.
    
\subsection{SVG-Net}

\begin{figure*}[h]
\begin{center}
   \includegraphics[width=1\linewidth]{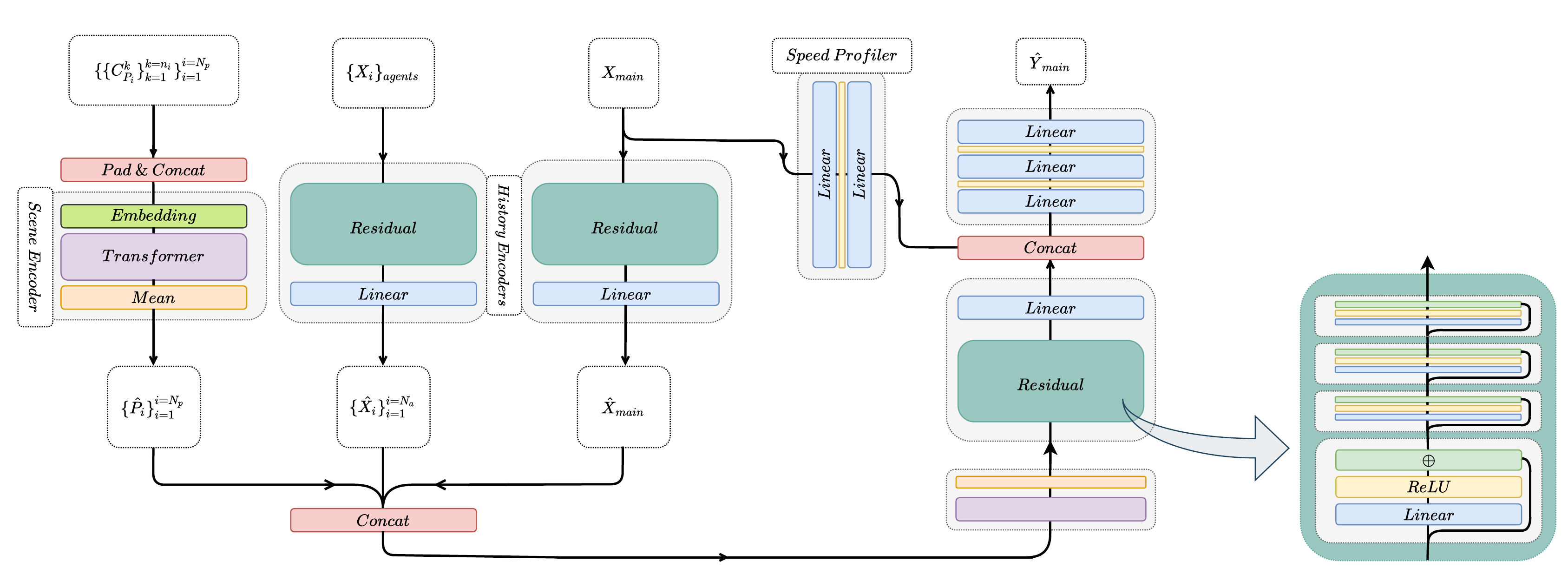}
\end{center}
   \caption{Overall structure of SVG-Net. The scene in the SVG format is processed by the transformer-based scene encoder. The histories of the main agent as well as other agents are encoded by History encoders. The learned latent representations are concatenated and passed to the next transformer to extract relations between agents and the scene. The learned features are then processed by a residual network and concatenated to main agent's history features. 
   The prediction would be the output of the final MLP given the concatenated features.}
\label{fig:network}
\end{figure*}

Inspired by \cite{carlier2020deepsvg}, to exploit SVG's hierarchical nature, a hierarchical network architecture, named SVG-Net, is employed. SVG-Net has three encoders and a single decoder. As Figure \ref{fig:network} depicts the block diagram of our SVG-Net's overall architecture, first, the encoders process the scene information, the history of the main agent, and the history of other agents in the scene into latent vectors, respectively. Then, the decoder extracts the relations between each latent representation to forecast the future of the main agent.

\noindent\textbf{Scene encoder.} To perceive the scene's information, the model should grasp each of its parts and make sense of the connections between them. The scene's contextual information is represented by a set of path objects, each defined by successive commands and represents parts of a lane or other infrastructures. The scene encoder is in charge of converting each path object to a latent representation, comprehensible to the decoder to extract the essential relations between them to recognize the scene eventually. Formally,
\begin{equation}
\begin{split}
    &\{\hat{P}_i\}_{i=1}^{i=N_P} = \{f_{scene}(P_i)\}_{i=1}^{i=N_P}, \\ 
    & \hat{P} = \hat{P}_1 \oplus ... \oplus \hat{P}_{N_p},
\end{split}
\end{equation}
explains the scene encoder's operation where $f_{scene}$ is the encoder function, $N_P$ is the number of paths, and $\oplus$ denotes concatenation. The scene encoder seeks to combine the information of separate SVG commands of each path and sight overall meaning of its described shape.
As suggested by the authors in \cite{carlier2020deepsvg} we employ a Transformer network \cite{vaswani2017attention} for this task since it is capable of looking at separate pairs of commands through its self-attention mechanism and combining them into higher level abstractions.

\noindent\textbf{History encoders.} The history encoders encode the preceding trajectories of the main agent and other agents in the scene. These encoders perceive each agent's motion state and their positional information and embed them into latent vectors. The decoder will then use this information to attend to the relative agents. Formally,  
\begin{equation}
\begin{split}
    &\{\hat{X}_i\}_1^{N_a} = \{f_{Hist}(X_i)\}_1^{N_a}, \\
    &f_{Hist}(.) = Lin(Residual_4(.)), \\ 
    & \hat{X}_{main} = \hat{X}_1, \\
    & \hat{X}_{agents} = \hat{X}_2 \oplus ... \oplus \hat{X}_{N_a},
\end{split}
\end{equation}
describes the history encoder's operation where $N_a$ denotes the number of agents in the scene, $Lin$ is a linear transformation, and $Residual_4$ is four layers of Multilayer Perceptron (MLP) residual network with ReLU non-linearities. 
We will report the performance of other alternatives in Section \ref{sec:ablation}.

\noindent\textbf{Decoder.}
The decoder takes the concatenated latent representations as input. Since it has to perceive the relations between the latents, we first employ a Transformer network to benefit from its powerful self-attention mechanism. The Transformer is followed by a residual MLP network to further process its findings in a comprehensible way to the final MLP module. We experimentally noted that providing a representation of the main agent's history to the final MLP is beneficial. Therefore, we employed an MLP network, called speed profiler to improve the prediction.
The final MLP will convert the processed latents to the predicted coordinates as follows:
\begin{equation}
\begin{split}
    &R = f_{Dec}(\hat{P} \oplus \hat{X}_{agents}  \oplus \hat{X}_{main}), \\
    &\hat{Y}_{main} =MLP( Residual_3(R) \oplus MLP(X_{main})),
\end{split}
\end{equation}
where $f_{Dec}$ is a Transformer followed by four layers of residual MLP network. 

\subsection{Loss function}
Our whole network is differentiable and can be trained in an end-to-end fashion. We leveraged Mean Squared Error (MSE) loss to supervise the network according to the following function:
\begin{equation}
    Loss_{MSE} = \sum_{t=T_{obs}+1}^{T_{obs}+T_{pred}}(\hat{Y}_{main}^t - Y_{main}^t)^2.
\end{equation}

\section{Experiments}
\subsection{Implementation details} 

The models are trained for 20 epochs with a batch size of 32. We employed AdamW optimizer \cite{DBLP:journals/corr/abs-1711-05101} with the initial learning rate of 0.0001, which is decreased by 0.9 every 2.5 epochs. 
As shown in Figure \ref{fig:network}, history encoders are composed of $4$  Linear residual blocks.
The transformer encoder and decoder consist of $4$ layers and $8$ self-attention heads.
In the end, the MLP network used as the speed profiler, and the final MLP network are $3$ and $2$ layer perceptrons, which have $(64)$ and $(128, 128)$ hidden layers, respectively. All the activation layers used in the model are ReLU layers.

\subsection{Dataset and metrics} 
Argoverse motion forecasting dataset \cite{chang2019argoverse} is a public dataset which consists of trajectory sequences recorded from two cities in Pennsylvania and Florida. 
The data holds a Vector Map of Lane Geometry that represents semantic road data as a localized graph and includes the centerlines of the street roads in the satellite map. 
The task is to forecast the next three seconds of each vehicle given the first two seconds of the motion history of the agents in the scene and the spatial context.
The data includes three disjoint training, validation, and test sets.

Our evaluation metrics are the common Average Displacement Error (ADE) and Final Displacement Error (FDE) in meters. We also report Miss rate (MR) which is the ratio of predictions whose final location is more than 2.0 meters away from the ground truth.
\subsubsection{Baselines}
We compare SVG-Net against the state-of-the-art methods on Argoverse test set. We report the two best baselines from \cite{chang2019argoverse} as simple but efficient methods: ``Constant Velocity" and ``LSTM ED". 
The next three baselines leverage vectorized scene representations.
We call them model-specific approaches as they all have their own way of processing the input.

The winner of Argo AI challenge in 2020,
``jean" \cite{argochall_2020,jean2020multi}, represents the lanes as Polylines and encodes them by 1D convolution and LSTM. They learn the interaction between agents and lanes with a transformer network. 

``VectorNet" \cite{gao2020_waymo_vectornet} builds a graph where each vector is treated as a node as well as trajectories of agents in the scene. Leveraging graph neural networks, they learn the relations between nodes and the final prediction. ``LaneGCN" \cite{liang2020_uber_learning} constructs a lane graph from vectorized scene and learns interactions between the lanes and the agents. 

The rasterized-based representation of scene is employed in ``uulm-mrm", the winner of AgoAI challenge in 2019 and one of the best methods with such standard scene representation.
``uulm-mrm" \cite{argochall_2019} renders the scene information as well as all agents' states on the image and employs a CNN to implicitly learn the interactions between the agents and the scene. 

\subsection{Results}
The performance of the model compared with state-of-the-art methods on the Argoverse test set is reported in Table \ref{tab:results}. In standardized format scene representations, we outperformed ``uulm-mrm" in all the three metrics. We hypothesize that it is both because we explicitly learn the relationships between different input components and our effective use of a dense input representation instead of utilizing sparse raster images. Compared with the models with more model-specific representations, our model has competitive performance. Note that the gap in the performance is the cost of using a standard representation which can be mitigated by introduction of more advanced methods in the SVG field.

\begin{table}
\begin{center}
\begin{tabular}{|l|l|c|c|c|}
\hline
Scene & \multirow{2}{*}{Model} & \multicolumn{3}{c|}{Argoverse-test}\\
Representation & & MR & FDE & ADE\\
\hline\hline
\multirow{2}{*}{Blind} & Constant velocity \cite{chang2019argoverse} & - & 7.89 & 3.55 \\
& LSTM ED \cite{chang2019argoverse}  & - & 4.67 & 2.25 \\
\hline
\multirow{3}{*}{Model-specific} 
& jean \cite{jean2020multi,argochall_2020} &  \textbf{0.59} & \textbf{3.73} & \textbf{1.68} \\
& VectorNet \cite{gao2020_waymo_vectornet}  & - & 4.01 & 1.81 \\
& LaneGCN \cite{liang2020_uber_learning}&  \textbf{0.59} & 3.78 & 1.71  \\
\hline
\multirow{2}{*}{Standard format} & uulm-mrm 
\cite{cui2019multimodal,argochall_2019} 
& 0.63 & 4.32 & 1.90 \\
& Ours & 0.62 & 3.96 & 1.80 \\ 
\hline
\end{tabular}
\end{center}
\captionsetup{font=small}
\caption{Comparing with state-of-the-art methods on Argoverse motion forecasting challenge (test set). For all metrics the lower is the better.}
\label{tab:results}
\end{table}

\subsection{Ablation study}\label{sec:ablation}
We perform ablation study to expose the impact of different building blocks of SVG-Net. 
\subsubsection{Impact of different inputs}
First, we study the influence of each input information on the final prediction. The results in Table \ref{tab:ablation_inputs} show that the scene information plays an important role in a correct prediction. This reveals the effectiveness of SVG-Net in understanding the scene from the SVG representation.
Incorporating agents further helps the model predict accurately which indicates that SVG-Net can learn the interactions among agents.

\begin{table}
\begin{center}
\begin{tabular}{|l|c|c|}
\hline
\multirow{2}{*}{Inputs} & \multicolumn{2}{c|}{Argoverse-test}\\
 & FDE & ADE  \\
\hline\hline
Hist &5.1 &2.25  \\
Hist + Scene  &4.4 &1.997 \\
Hist + Scene + Agents & \textbf{3.96} &\textbf{1.80} \\ 
\hline
\end{tabular}
\end{center}
\captionsetup{font=small}
\caption{Ablation study on the impact of each input on the final performance. 
}
\label{tab:ablation_inputs}
\end{table}

\subsubsection{SVG against raster images}
Then, we study how the use of SVG representation is beneficial compared to raster images when applied to the same architecture. 

We replaced the SVG inputs with rasterized images and the scene encoder with a Resnet-18 \cite{he2016deep} while the rest of the SVG-Net network is not changed. The first row in Table \ref{tab:ablation_model} shows that using SVG-based model outperforms the use of rasterized inputs. This indicates that while understanding the scene from the rasterized high-dimensional data is possible, the densely vectorized nature of SVG data allows the model to better understand the scene.

\subsubsection{Alternatives for model networks} 

Finally, in order to identify the best choices for the encoders and the decoder architecture, we examined multiple alternatives in Table \ref{tab:ablation_model}. 
The second row implies the use of transformer network instead of the residual history encoders. 
In another experiment, shown in the third row of the table, we replaced the transformers in scene encoder with residual MLPs. Based on the results of these two experiments, we observed that transformers are able to better learn the shapes while the residual MLPs are better in embedding the motion state and positional information.\\
Motivated by recent progress in language modeling, we also tried utilizing the recent network architecture, Albert \cite{lan2019albert} which performed worse than our bare transformer network.

\begin{table}
\begin{center}
\begin{tabular}{|l|c|c|} 
\hline
\multirow{2}{*}{Alternative models} & \multicolumn{2}{c|}{Argoverse-test}\\
  & FDE & ADE \\
\hline\hline
Rasterized scene & 4.35 & 1.95  \\
Transformer encoders & 4.37 & 1.97  \\
MLP encoded paths & 6.34 &2.75  \\
Albert network & 4.15 & 1.87\\
SVG-Net &\textbf{3.96} &\textbf{1.80}  \\
\hline
\end{tabular}
\end{center}
\captionsetup{font=small}
\caption{Studying the impact of using different alternatives for each part of SVG-Net network on the performance.}
\label{tab:ablation_model}
\end{table}
\subsection{Advantages of using SVG} 
In order to experimentally expose the benefits of using a standard representation, we conduct the following experiments:
\subsubsection{Leveraging other SVG data}
Thanks to having a standard input representation, we studied if we can leverage the learned representation from other datasets and potentially other tasks to the benefit of our own problem.
To this end, we pick the reconstruction task on the SVG-Icons8 dataset \cite{carlier2020deepsvg}. First, SVG-Net encoder is trained with the decoder of \cite{carlier2020deepsvg} to perform the reconstruction. Then, we freeze the scene encoder and fine-tune the rest of SVG-Net. Note that the scene encoder is frozen to better assess the generalizability of the learned representation. 
The results are reported in Table \ref{tab:transfer}.
Impressively, the transferred model has achieved a close performance with the original model. 
This indicates that its representation could generalize to the new task of prediction. While these results are achieved using the limited SVG-Icons8 dataset, we hypothesize that having larger and more diverse datasets, similar to Imagenet \cite{deng2009imagenet}, will definitely be more impactful.

\begin{table}
\begin{center}
\begin{tabular}{|l|c|c|} 
\hline
\multirow{2}{*}{Model} &\multicolumn{2}{c|}{Argoverse-val}\\
  & FDE & ADE \\
\hline\hline
SVG-Net & 3.25 & 1.46  \\
Pre-trained on SVG-Icons8 dataset & 3.42 & 1.51  \\
\hline
\end{tabular}
\end{center}
\captionsetup{font=small}
\caption{Demonstrating the possibility of using other SVG data for pre-training the network.
}
\label{tab:transfer}
\end{table}

\begin{table}
\begin{center}
\begin{tabular}{|l|c|} 
\hline
Model & Accuracy \\
\hline\hline
SVG-Net & 72.2 \%  \\
Resnet18 \cite{he2016deep} & 73.0 \% \\
\hline
\end{tabular}
\end{center}
\captionsetup{font=small}
\caption{Demonstrating the possibility of using SVG-Net network for the classification task over SVG-Icons8 dataset.
}
\label{tab:classification}
\end{table}

\subsubsection{Knowledge transfer}
The advancements in CNN architectures have been the workhorse of progress in many computer vision tasks. This is due to the fact that CNNs are used to learn a representation from standard pixel-based (rasterized) representation of images. Therefore, any advancements in CNN structure can be beneficial for other models. Similarly, replacing rasterized data with the standard SVG format would give the chance to other models with SVG inputs to leverage the advancements for learning representation of inputs. We demonstrate this by showing the effectiveness of our method on the classification task on the Font dataset \cite{fontdataset}. We keep the SVG-Net architecture, remove the agents from inputs and change the output to predict classes. The results are shown in table \ref{tab:classification} where the method performs on par with a residual CNN \cite{he2016deep}.

\begin{figure}[t]
\begin{center}
\centering
 \vspace{9mm}
 \subfloat{\includegraphics[width=0.47\linewidth]{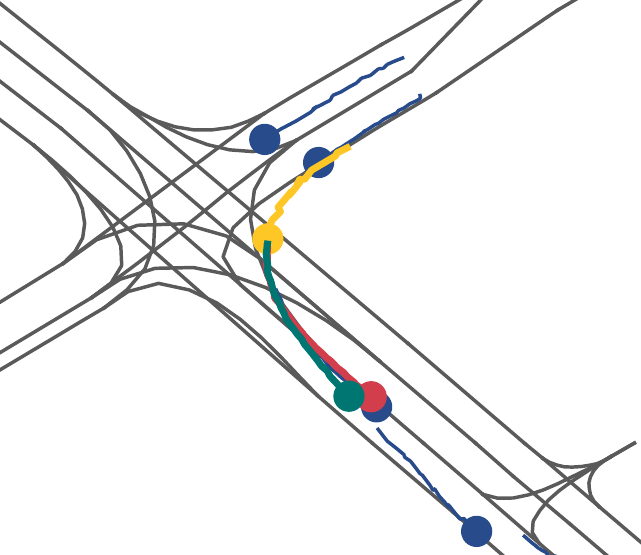}
}
\subfloat{\includegraphics[width=0.47\linewidth]{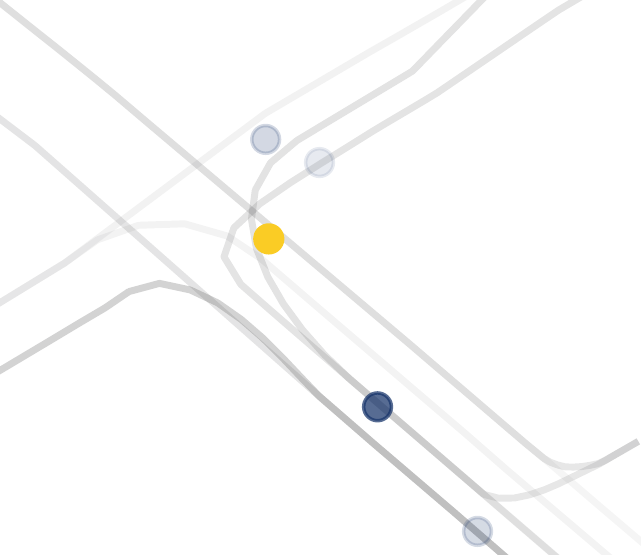}
}
\\
\vspace{9mm}
\subfloat{\includegraphics[width=0.47\linewidth]{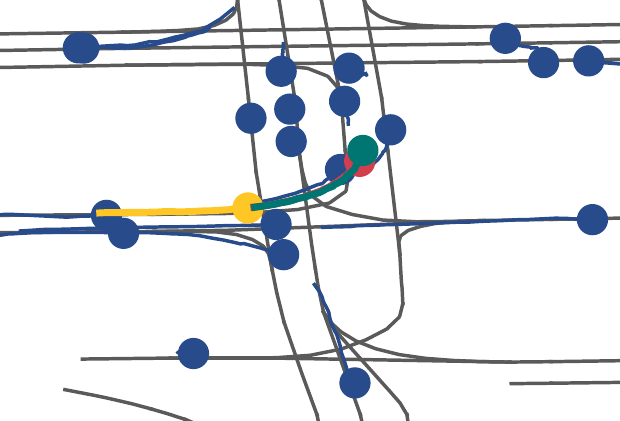}
}
\subfloat{\includegraphics[width=0.47\linewidth]{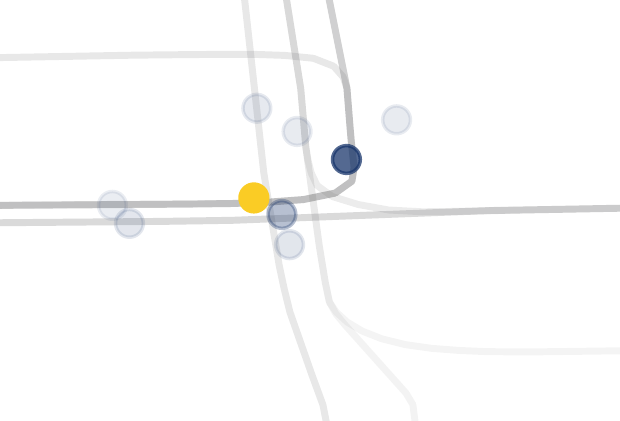}
}
\\
 \vspace{9mm}
\hfill
\subfloat{\includegraphics[width=0.43\linewidth]{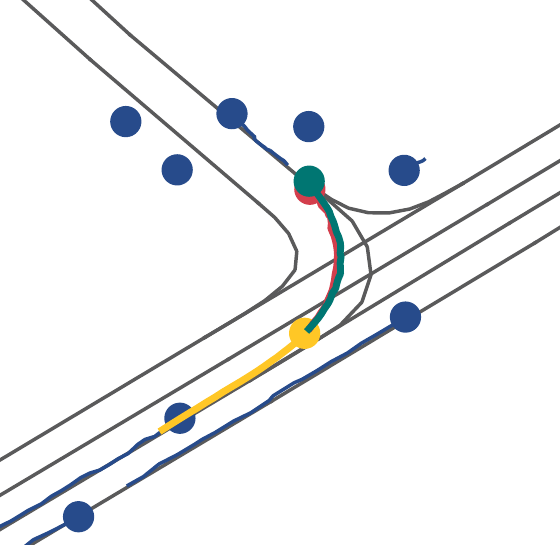}
}
\subfloat{\includegraphics[width=0.43\linewidth]{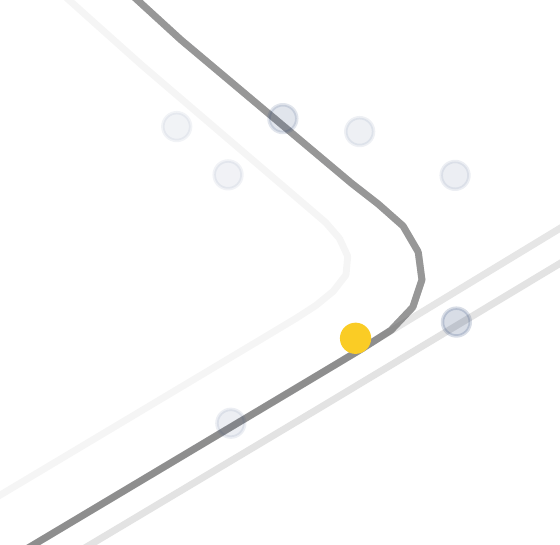}
}

\end{center}
\vspace{-15mm}
\hspace{26mm}
\subfloat{\includegraphics[width=0.4\linewidth]{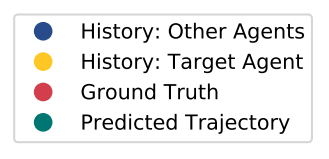}
}
\caption{Qualitative results of the predictions. On the left, the predictions of the model are visualized and on the right, the attended regions and agents are shown. 
We visualize higher attention values with more opacity. The first image shows the ability of the model in perceiving the SVG-based scene. The second scenario demonstrates the ability of the network in identifying interacting agents. The last case shows understanding of the model of the agent's goal based on its history which helped attending to the right lane. 
}
\label{fig:qualitative}
\end{figure}

\subsection{Qualitative results}
Figure \ref{fig:qualitative} shows the performance of the model qualitatively. In the left column, predictions of the model are depicted in different scenarios. 
In the right column, the attention of the transformer decoder with respect to the main agent being queried is highlighted. The higher the color's opacity of a component, the higher is the attention score of the main agent's history for that particular component.  The results demonstrate that the model can successfully perceive the SVG-represented scene as well as interactions with other agents. It can be seen that attention scores match our expectations. The model could detect the right lanes. It also flawlessly identifies the interacting agents. When multiple plausible options are available, the attention correctly matches with both cases. In the last scenario, even the model attends partially to the agent on the non-taken road.

\section{Conclusions}
We have presented a new input format and representation to solve the vehicle trajectory prediction task. Instead of using rasterized or model-specific vectorized representations, we propose using the well-known standard SVG format for representing the scene context. To effectively perceive the scene as well as interactions among agents, we offer SVG-Net, an encoder-decoder model based on the transformer network. Our transformer-based scene encoder understands the scene context. Also, the history encoders capture the motion history employing residual MLP. The transformer-based decoder extracts the relations between the learned embeddings. Conducting experiments on Argoverse public dataset, we demonstrate the effectiveness of SVG-Net. We also assess the importance of different blocks in the model 
and expose the advantages of using a standard data format.

We hope that our study encourages the community to consider SVG as a research-worthy straightforward representation of scene context for the trajectory prediction task. As future work, we will continue to investigate and leverage new network architectures to best model SVG inputs. We believe that by using a standard vectorized format, we will be able to take advantage of related works in the field dealing with such inputs and reproduce the same success stories we witnessed with images.

{\small
\bibliographystyle{ieee_fullname}
\bibliography{egbib}

\begin{thebibliography}{10}\itemsep=-1pt

\bibitem{argochall_2019}
Argoai challenge 2019. neurips workshop on machine learning for autonomous
  driving (2019).
\newblock \url{https://slideslive.com/38923162/argoai-challenge}.

\bibitem{argochall_2020}
Argoai challenge 2020. cvpr workshop on autonomous driving (2019).
\newblock \url{https://www.youtube.com/watch?v=Vcbj\_peZT4Q}.

\bibitem{w3c_svg}
The w3c svg working group.
\newblock \url{https://www.w3.org/Graphics/SVG}.

\bibitem{waymo_challenge}
Waymo open dataset motion prediction challenge, 2021.
\newblock
  \url{https://waymo.com/intl/es\_ALL/open/challenges/2021/motion-prediction/}.

\bibitem{alahi2016sociallstm}
Alexandre Alahi, Kratarth Goel, Vignesh Ramanathan, Alexandre Robicquet, Li
  Fei-Fei, and Silvio Savarese.
\newblock Social lstm: Human trajectory prediction in crowded spaces.
\newblock In {\em Proceedings of the IEEE conference on computer vision and
  pattern recognition}, pages 961--971, 2016.

\bibitem{alahi2016social}
Alexandre Alahi, Kratarth Goel, Vignesh Ramanathan, Alexandre Robicquet, Li
  Fei-Fei, and Silvio Savarese.
\newblock Social lstm: Human trajectory prediction in crowded spaces.
\newblock In {\em Proceedings of the IEEE conference on computer vision and
  pattern recognition}, pages 961--971, 2016.

\bibitem{Chauffeurnet_waymo}
Mayank Bansal, Alex Krizhevsky, and Abhijit Ogale.
\newblock Chauffeurnet: Learning to drive by imitating the best and
  synthesizing the worst.
\newblock {\em arXiv preprint arXiv:1812.03079}, 2018.

\bibitem{bartoli2018context}
Federico Bartoli, Giuseppe Lisanti, Lamberto Ballan, and Alberto Del~Bimbo.
\newblock Context-aware trajectory prediction.
\newblock In {\em 2018 24th International Conference on Pattern Recognition
  (ICPR)}, pages 1941--1946. IEEE, 2018.

\bibitem{nuscenes2019}
Holger Caesar, Varun Bankiti, Alex~H. Lang, Sourabh Vora, Venice~Erin Liong,
  Qiang Xu, Anush Krishnan, Yu Pan, Giancarlo Baldan, and Oscar Beijbom.
\newblock nuscenes: A multimodal dataset for autonomous driving.
\newblock {\em arXiv preprint arXiv:1903.11027}, 2019.

\bibitem{carlier2020deepsvg}
Alexandre Carlier, Martin Danelljan, Alexandre Alahi, and Radu Timofte.
\newblock Deepsvg: A hierarchical generative network for vector graphics
  animation.
\newblock {\em Advances in Neural Information Processing Systems}, 33, 2020.

\bibitem{chai2019multipath}
Yuning Chai, Benjamin Sapp, Mayank Bansal, and Dragomir Anguelov.
\newblock Multipath: Multiple probabilistic anchor trajectory hypotheses for
  behavior prediction.
\newblock {\em arXiv preprint arXiv:1910.05449}, 2019.

\bibitem{chang2019argoverse}
Ming-Fang Chang, John Lambert, Patsorn Sangkloy, Jagjeet Singh, Slawomir Bak,
  Andrew Hartnett, De Wang, Peter Carr, Simon Lucey, Deva Ramanan, et~al.
\newblock Argoverse: 3d tracking and forecasting with rich maps.
\newblock In {\em Proceedings of the IEEE Conference on Computer Vision and
  Pattern Recognition}, pages 8748--8757, 2019.

\bibitem{Argoverse}
Ming-Fang Chang, John~W Lambert, Patsorn Sangkloy, Jagjeet Singh, Slawomir Bak,
  Andrew Hartnett, De Wang, Peter Carr, Simon Lucey, Deva Ramanan, and James
  Hays.
\newblock Argoverse: 3d tracking and forecasting with rich maps.
\newblock In {\em Conference on Computer Vision and Pattern Recognition
  (CVPR)}, 2019.

\bibitem{cui2019multimodal}
Henggang Cui, Vladan Radosavljevic, Fang-Chieh Chou, Tsung-Han Lin, Thi Nguyen,
  Tzu-Kuo Huang, Jeff Schneider, and Nemanja Djuric.
\newblock Multimodal trajectory predictions for autonomous driving using deep
  convolutional networks.
\newblock In {\em 2019 International Conference on Robotics and Automation
  (ICRA)}, pages 2090--2096. IEEE, 2019.

\bibitem{das2021cloud2curve}
Ayan Das, Yongxin Yang, Timothy~M Hospedales, Tao Xiang, and Yi-Zhe Song.
\newblock Cloud2curve: Generation and vectorization of parametric sketches.
\newblock In {\em Proceedings of the IEEE/CVF Conference on Computer Vision and
  Pattern Recognition}, pages 7088--7097, 2021.

\bibitem{deng2009imagenet}
Jia Deng, Wei Dong, Richard Socher, Li-Jia Li, Kai Li, and Li Fei-Fei.
\newblock Imagenet: A large-scale hierarchical image database.
\newblock In {\em 2009 IEEE conference on computer vision and pattern
  recognition}, pages 248--255. Ieee, 2009.

\bibitem{egiazarian2020deepVectorization}
Vage Egiazarian, Oleg Voynov, Alexey Artemov, Denis Volkhonskiy, Aleksandr
  Safin, Maria Taktasheva, Denis Zorin, and Evgeny Burnaev.
\newblock Deep vectorization of technical drawings.
\newblock In {\em European Conference on Computer Vision}, pages 582--598.
  Springer, 2020.

\bibitem{gao2020_waymo_vectornet}
Jiyang Gao, Chen Sun, Hang Zhao, Yi Shen, Dragomir Anguelov, Congcong Li, and
  Cordelia Schmid.
\newblock Vectornet: Encoding hd maps and agent dynamics from vectorized
  representation.
\newblock In {\em Proceedings of the IEEE/CVF Conference on Computer Vision and
  Pattern Recognition}, pages 11525--11533, 2020.

\bibitem{girshick2015fast}
Ross Girshick.
\newblock Fast r-cnn.
\newblock In {\em Proceedings of the IEEE international conference on computer
  vision}, pages 1440--1448, 2015.

\bibitem{guo2019deepline}
Yi Guo, Zhuming Zhang, Chu Han, Wenbo Hu, Chengze Li, and Tien-Tsin Wong.
\newblock Deep line drawing vectorization via line subdivision and topology
  reconstruction.
\newblock In {\em Computer Graphics Forum}, volume~38, pages 81--90. Wiley
  Online Library, 2019.

\bibitem{he2016deep}
Kaiming He, Xiangyu Zhang, Shaoqing Ren, and Jian Sun.
\newblock Deep residual learning for image recognition.
\newblock In {\em Proceedings of the IEEE conference on computer vision and
  pattern recognition}, pages 770--778, 2016.

\bibitem{social-forces}
Dirk Helbing and Peter Molnar.
\newblock Social force model for pedestrian dynamics.
\newblock {\em Physical Review E}, 51, 05 1998.

\bibitem{hong2019rules}
Joey Hong, Benjamin Sapp, and James Philbin.
\newblock Rules of the road: Predicting driving behavior with a convolutional
  model of semantic interactions.
\newblock In {\em Proceedings of the IEEE Conference on Computer Vision and
  Pattern Recognition}, pages 8454--8462, 2019.

\bibitem{lyft2020}
John Houston, Guido Zuidhof, Luca Bergamini, Yawei Ye, Ashesh Jain, Sammy
  Omari, Vladimir Iglovikov, and Peter Ondruska.
\newblock One thousand and one hours: Self-driving motion prediction dataset,
  2020.

\bibitem{frenet}
Y. {Hu}, L. {Sun}, and M. {Tomizuka}.
\newblock Generic prediction architecture considering both rational and
  irrational driving behaviors.
\newblock In {\em 2019 IEEE Intelligent Transportation Systems Conference
  (ITSC)}, pages 3539--3546, 2019.

\bibitem{huang2017densely}
Gao Huang, Zhuang Liu, Laurens Van Der~Maaten, and Kilian~Q Weinberger.
\newblock Densely connected convolutional networks.
\newblock In {\em Proceedings of the IEEE conference on computer vision and
  pattern recognition}, pages 4700--4708, 2017.

\bibitem{ivanovic2019trajectron}
Boris Ivanovic and Marco Pavone.
\newblock The trajectron: Probabilistic multi-agent trajectory modeling with
  dynamic spatiotemporal graphs.
\newblock In {\em Proceedings of the IEEE International Conference on Computer
  Vision}, pages 2375--2384, 2019.

\bibitem{curbside}
N. {Jaipuria}, G. {Habibi}, and J.~P. {How}.
\newblock Learning in the curbside coordinate frame for a transferable
  pedestrian trajectory prediction model.
\newblock In {\em 2018 21st International Conference on Intelligent
  Transportation Systems (ITSC)}, pages 3125--3131, 2018.

\bibitem{kingma2013auto}
Diederik~P Kingma and Max Welling.
\newblock Auto-encoding variational bayes.
\newblock {\em arXiv preprint arXiv:1312.6114}, 2013.

\bibitem{kothari2020human}
Parth Kothari, Sven Kreiss, and Alexandre Alahi.
\newblock Human trajectory forecasting in crowds: A deep learning perspective.
\newblock {\em arXiv preprint arXiv:2007.03639}, 2020.

\bibitem{lan2019albert}
Zhenzhong Lan, Mingda Chen, Sebastian Goodman, Kevin Gimpel, Piyush Sharma, and
  Radu Soricut.
\newblock Albert: A lite bert for self-supervised learning of language
  representations.
\newblock {\em arXiv preprint arXiv:1909.11942}, 2019.

\bibitem{li2020differentiable}
Tzu-Mao Li, Michal Luk{\'a}{\v{c}}, Micha{\"e}l Gharbi, and Jonathan
  Ragan-Kelley.
\newblock Differentiable vector graphics rasterization for editing and
  learning.
\newblock {\em ACM Transactions on Graphics (TOG)}, 39(6):1--15, 2020.

\bibitem{liang2020_uber_learning}
Ming Liang, Bin Yang, Rui Hu, Yun Chen, Renjie Liao, Song Feng, and Raquel
  Urtasun.
\newblock Learning lane graph representations for motion forecasting.
\newblock {\em arXiv preprint arXiv:2007.13732}, 2020.

\bibitem{liu2017raster}
Chen Liu, Jiajun Wu, Pushmeet Kohli, and Yasutaka Furukawa.
\newblock Raster-to-vector: Revisiting floorplan transformation.
\newblock In {\em Proceedings of the IEEE International Conference on Computer
  Vision}, pages 2195--2203, 2017.

\bibitem{lopes2019learned}
Raphael~Gontijo Lopes, David Ha, Douglas Eck, and Jonathon Shlens.
\newblock A learned representation for scalable vector graphics.
\newblock In {\em Proceedings of the IEEE International Conference on Computer
  Vision}, pages 7930--7939, 2019.

\bibitem{fontdataset}
Raphael~Gontijo Lopes, David Ha, Douglas Eck, and Jonathon Shlens.
\newblock A learned representation for scalable vector graphics.
\newblock In {\em 2019 IEEE/CVF International Conference on Computer Vision
  (ICCV)}, pages 7929--7938, 2019.

\bibitem{DBLP:journals/corr/abs-1711-05101}
Ilya Loshchilov and Frank Hutter.
\newblock Fixing weight decay regularization in adam.
\newblock {\em CoRR}, abs/1711.05101, 2017.

\bibitem{jean2020multi}
Jean Mercat, Thomas Gilles, Nicole El~Zoghby, Guillaume Sandou, Dominique
  Beauvois, and Guillermo~Pita Gil.
\newblock Multi-head attention for multi-modal joint vehicle motion
  forecasting.
\newblock In {\em 2020 IEEE International Conference on Robotics and Automation
  (ICRA)}, pages 9638--9644. IEEE, 2020.

\bibitem{park2020diverse}
Seong~Hyeon Park, Gyubok Lee, Jimin Seo, Manoj Bhat, Minseok Kang, Jonathan
  Francis, Ashwin Jadhav, Paul~Pu Liang, and Louis-Philippe Morency.
\newblock Diverse and admissible trajectory forecasting through multimodal
  context understanding.
\newblock In {\em European Conference on Computer Vision}, pages 282--298.
  Springer, 2020.

\bibitem{reddy2021im2vec}
Pradyumna Reddy, Michael Gharbi, Michal Lukac, and Niloy~J Mitra.
\newblock Im2vec: Synthesizing vector graphics without vector supervision.
\newblock {\em arXiv preprint arXiv:2102.02798}, 2021.

\bibitem{redmon2016you}
Joseph Redmon, Santosh Divvala, Ross Girshick, and Ali Farhadi.
\newblock You only look once: Unified, real-time object detection.
\newblock In {\em Proceedings of the IEEE conference on computer vision and
  pattern recognition}, pages 779--788, 2016.

\bibitem{ren2021safety}
Xuanchi Ren, Tao Yang, Li~Erran Li, Alexandre Alahi, and Qifeng Chen.
\newblock Safety-aware motion prediction with unseen vehicles for autonomous
  driving.
\newblock 2021.

\bibitem{saadatnejad2021socially}
Saeed Saadatnejad, Mohammadhossein Bahari, Pedram Khorsandi, Mohammad Saneian,
  Seyed-Mohsen Moosavi-Dezfooli, and Alexandre Alahi.
\newblock Are socially-aware trajectory prediction models really
  socially-aware?
\newblock {\em arXiv preprint arXiv:2108.10879}, 2021.

\bibitem{sadeghian2018car}
Amir Sadeghian, Ferdinand Legros, Maxime Voisin, Ricky Vesel, Alexandre Alahi,
  and Silvio Savarese.
\newblock Car-net: Clairvoyant attentive recurrent network.
\newblock In {\em Proceedings of the European Conference on Computer Vision
  (ECCV)}, pages 151--167, 2018.

\bibitem{salzmann2020trajectron}
Tim Salzmann, Boris Ivanovic, Punarjay Chakravarty, and Marco Pavone.
\newblock Trajectron++: Dynamically-feasible trajectory forecasting with
  heterogeneous data.
\newblock In {\em Computer Vision--ECCV 2020: 16th European Conference,
  Glasgow, UK, August 23--28, 2020, Proceedings, Part XVIII 16}, pages
  683--700. Springer, 2020.

\bibitem{shi2021sgcn}
Liushuai Shi, Le Wang, Chengjiang Long, Sanping Zhou, Mo Zhou, Zhenxing Niu,
  and Gang Hua.
\newblock Sgcn: Sparse graph convolution network for pedestrian trajectory
  prediction.
\newblock In {\em Proceedings of the IEEE/CVF Conference on Computer Vision and
  Pattern Recognition}, pages 8994--9003, 2021.

\bibitem{leader-follower}
Martin Treiber, Ansgar Hennecke, and Dirk Helbing.
\newblock Congested traffic states in empirical observations and microscopic
  simulations.
\newblock {\em Physical review E}, 62(2):1805, 2000.

\bibitem{vaswani2017attention}
Ashish Vaswani, Noam Shazeer, Niki Parmar, Jakob Uszkoreit, Llion Jones,
  Aidan~N Gomez, {\L}ukasz Kaiser, and Illia Polosukhin.
\newblock Attention is all you need.
\newblock In {\em Advances in neural information processing systems}, pages
  5998--6008, 2017.

\bibitem{vemula2018socialattention}
Anirudh Vemula, Katharina Muelling, and Jean Oh.
\newblock Social attention: Modeling attention in human crowds.
\newblock In {\em 2018 IEEE international Conference on Robotics and Automation
  (ICRA)}, pages 1--7. IEEE, 2018.

\bibitem{xie2020self}
Qizhe Xie, Minh-Thang Luong, Eduard Hovy, and Quoc~V Le.
\newblock Self-training with noisy student improves imagenet classification.
\newblock In {\em Proceedings of the IEEE/CVF Conference on Computer Vision and
  Pattern Recognition}, pages 10687--10698, 2020.

\bibitem{noticing2021}
Dapeng Zhao and Jean Oh.
\newblock Noticing motion patterns: A temporal cnn with a novel convolution
  operator for human trajectory prediction.
\newblock {\em IEEE Robotics and Automation Letters}, 6(2):628--634, 2021.

\bibitem{benz}
Julius Ziegler, Philipp Bender, Markus Schreiber, Henning Lategahn, Tobias
  Strauss, Christoph Stiller, Thao Dang, Uwe Franke, Nils Appenrodt,
  Christoph~G Keller, et~al.
\newblock Making bertha drive—an autonomous journey on a historic route.
\newblock {\em IEEE Intelligent transportation systems magazine}, 6(2):8--20,
  2014.

\end{thebibliography}
}

\section{Supplementary Material}
\subsection{Network Architecture Details}
\begin{figure*}[!h]
    \centering
    \includegraphics[width=\linewidth]{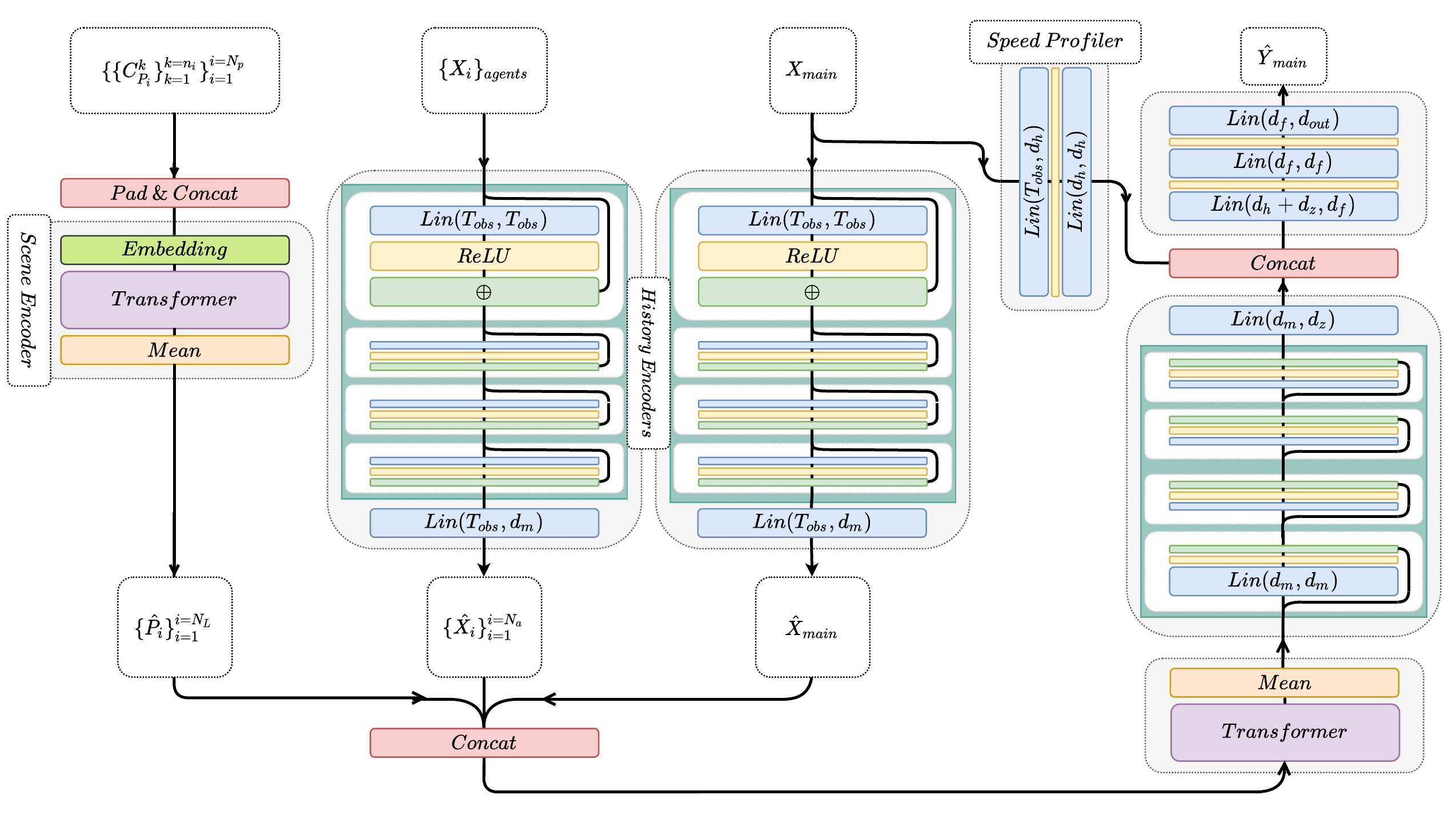}
    \caption{Architecture of SVG-Net. We provide more details on each block.}
    \label{fig:model-details}
\end{figure*}

Our model structure is shown in more details in Figure \ref{fig:model-details} where $T_{obs}=20, d_m=256, d_h=40, d_z=64, d_f=128, d_{out}=60$. We use ReLU as the non-linear activation function. The transformers have 4 layer with 8 attention heads. We use the same embeddings as \cite{carlier2020deepsvg} for the SVG commands.

\subsection{Extra results on Argoverse dataset:}
\begin{figure}[t]
\begin{center}
\centering
\vspace{1mm}
\subfloat{\includegraphics[width=0.47\linewidth]{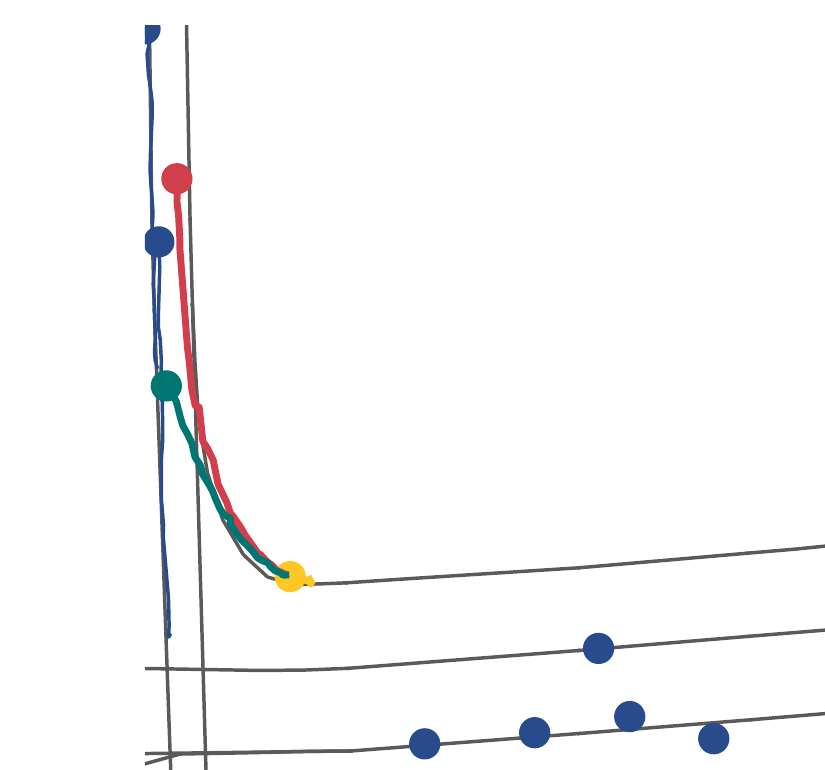}}
\subfloat{\includegraphics[width=0.47\linewidth]{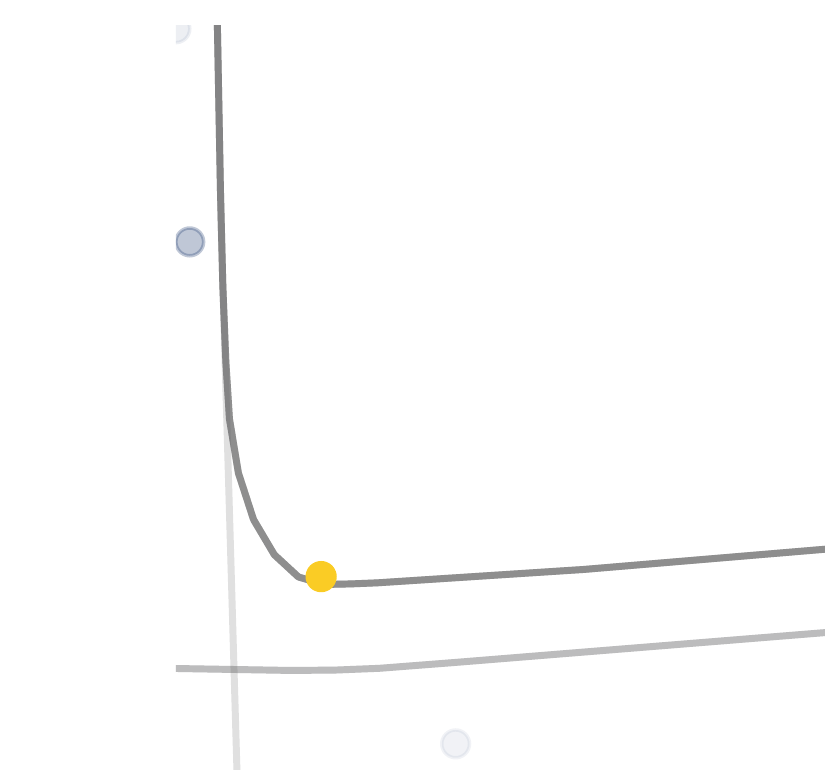}}\\
\vspace{1mm}
\subfloat{\includegraphics[width=0.47\linewidth]{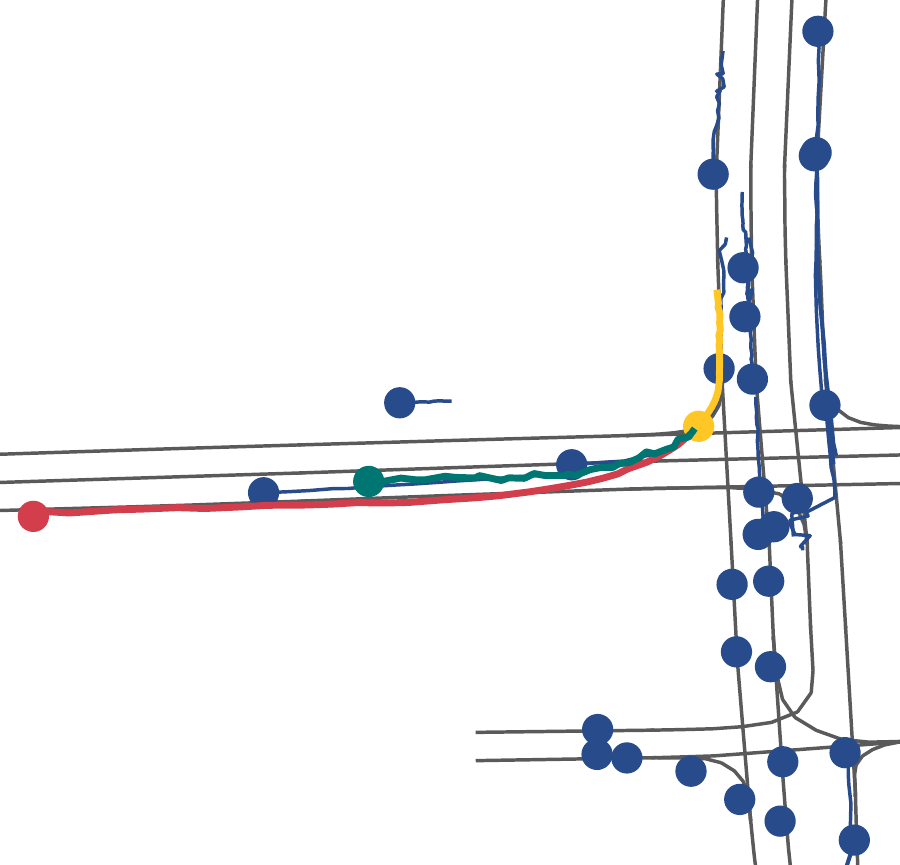}}
\subfloat{\includegraphics[width=0.47\linewidth]{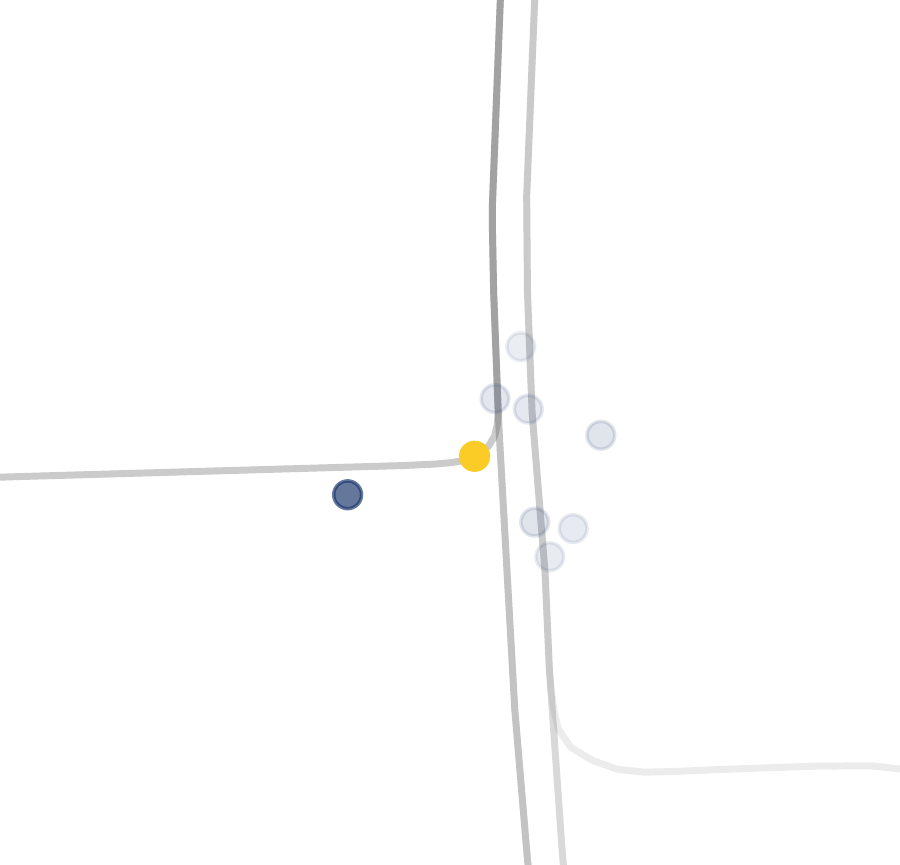}}\\
 \vspace{1mm}
\subfloat{\includegraphics[width=0.49\linewidth]{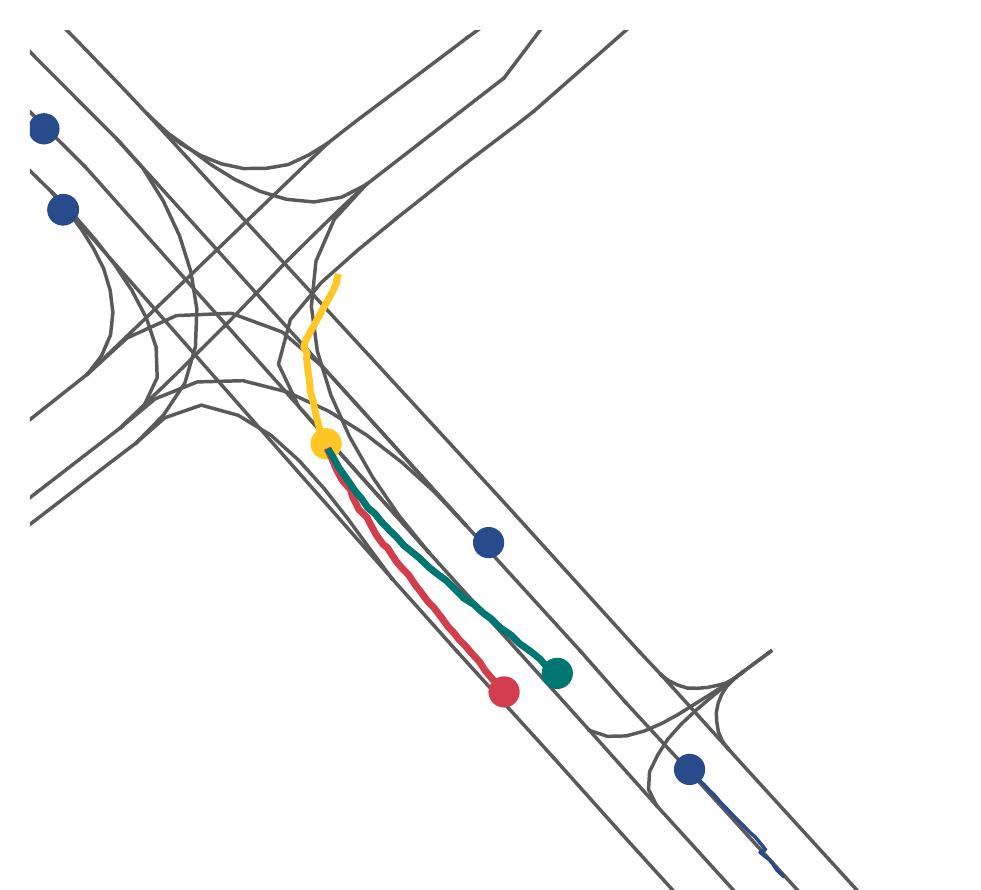}}
\subfloat{\includegraphics[width=0.49\linewidth]{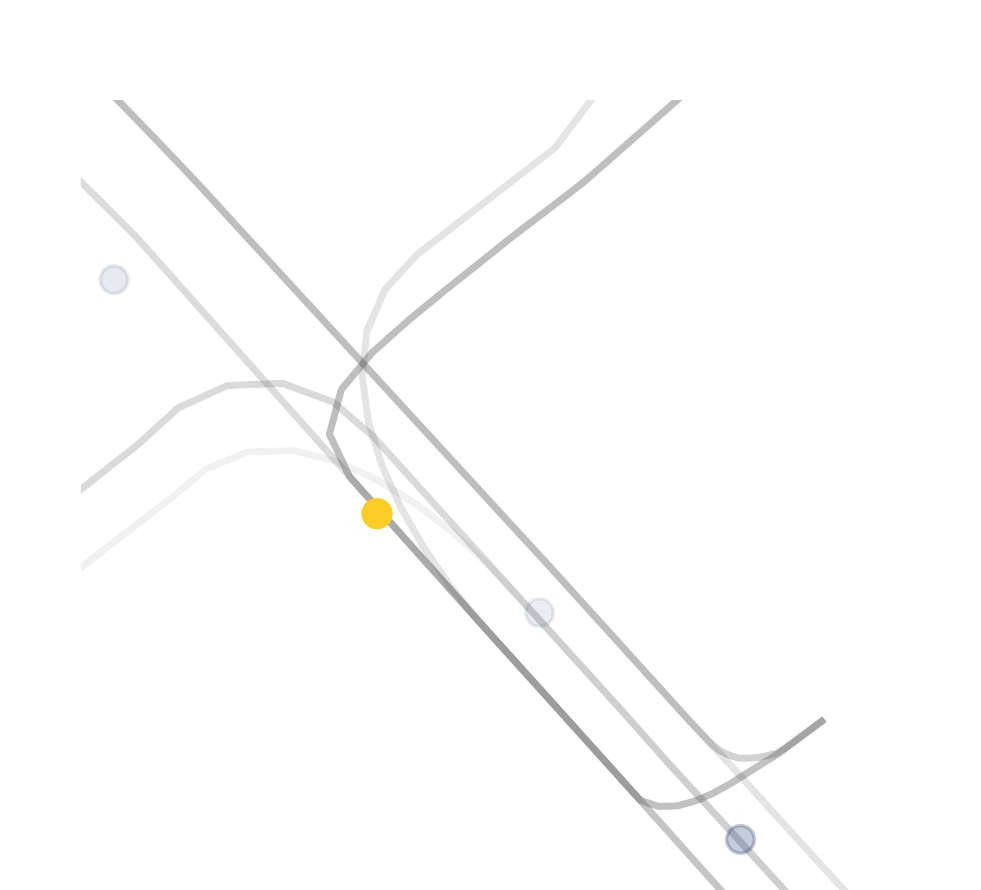}}\\
\vspace{1mm}
\subfloat{\includegraphics[width=0.48\linewidth]{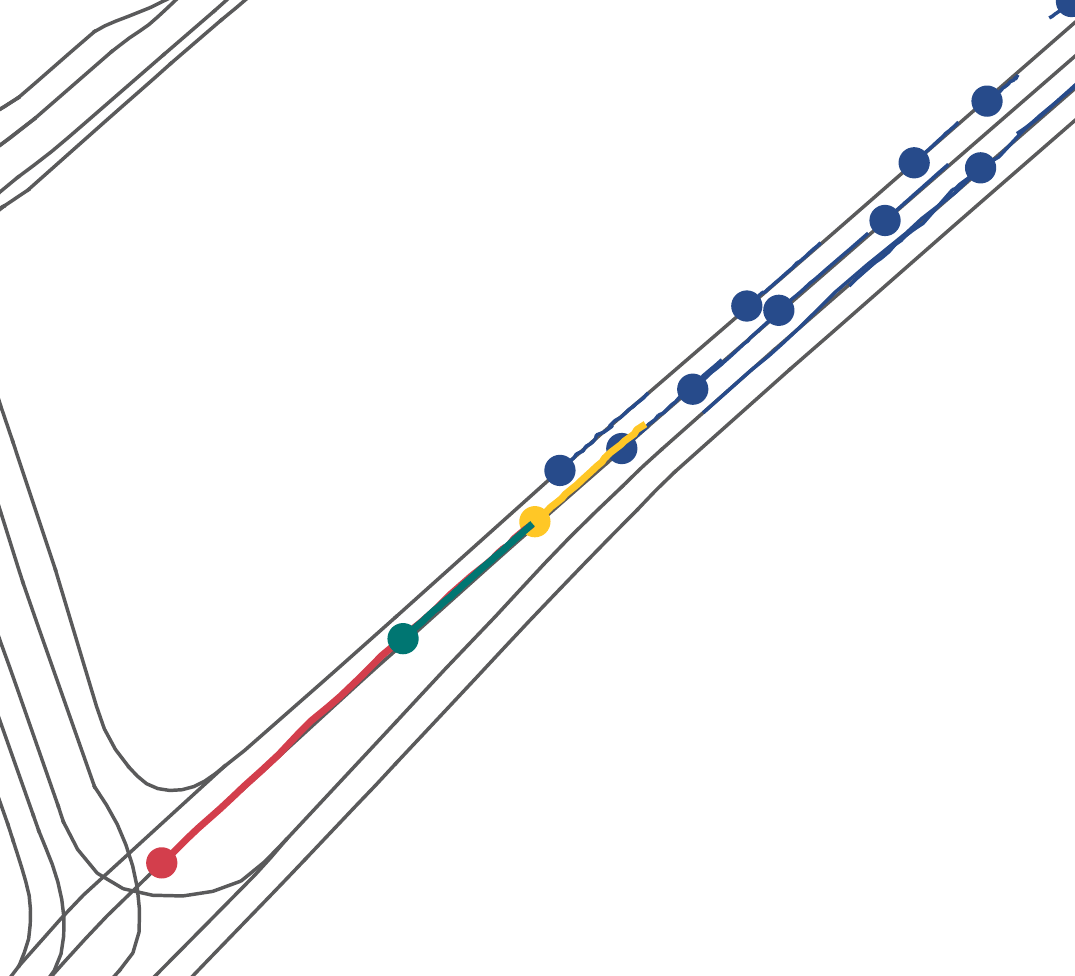}}
\subfloat{\includegraphics[width=0.48\linewidth]{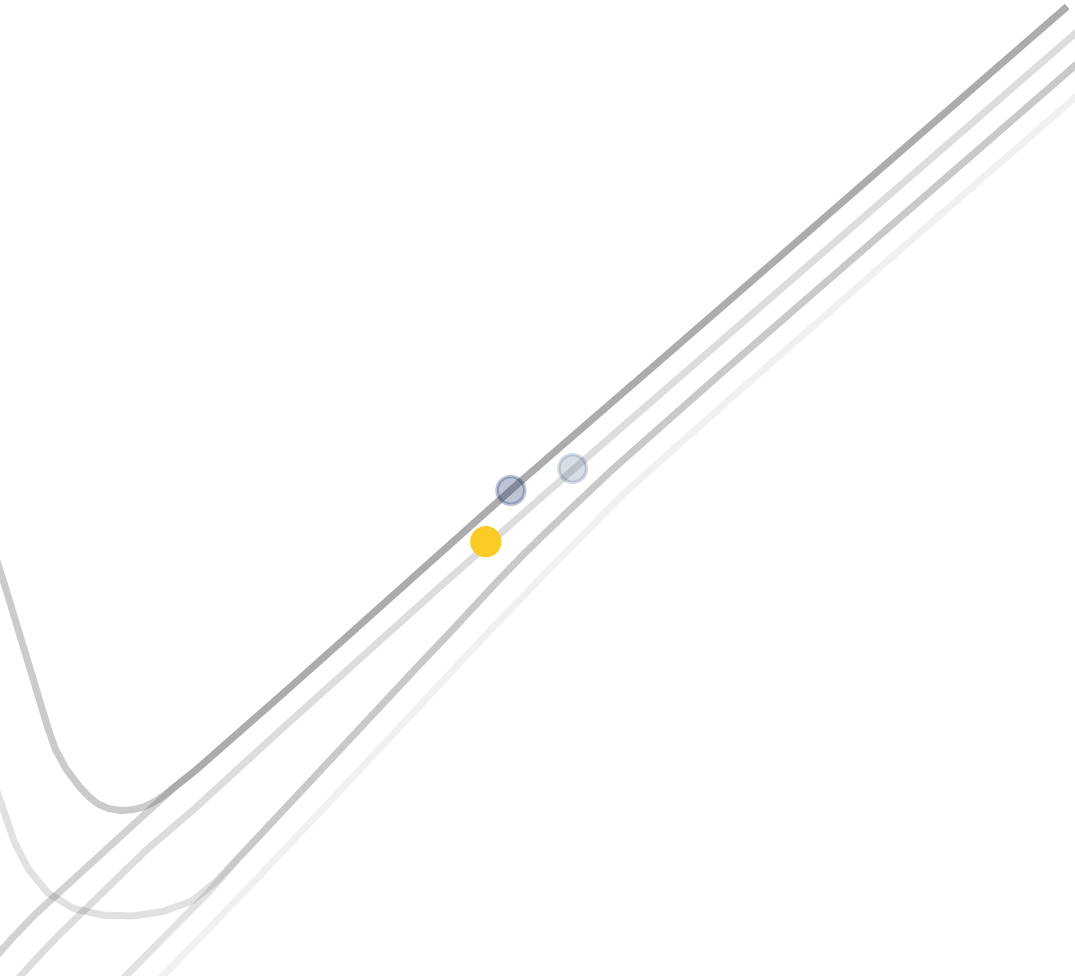}}
\end{center}
\vspace{-15mm}
\hspace{26mm}
\subfloat{\includegraphics[width=0.4\linewidth]{images-supp/vis-legend.PNG}
}
\caption{Some failure cases of the model on the left along with their corresponding visualization of attended regions on the right.}
\label{asghar}
\end{figure}

Figure \ref{fig:qualitative} shows more qualitative results on Argoverse dataset. In Figure \ref{asghar}, we provide some failure cases of the model.


\begin{figure}[t]
\begin{center}
\centering
 \vspace{1mm}
 \subfloat{\includegraphics[width=0.47\linewidth]{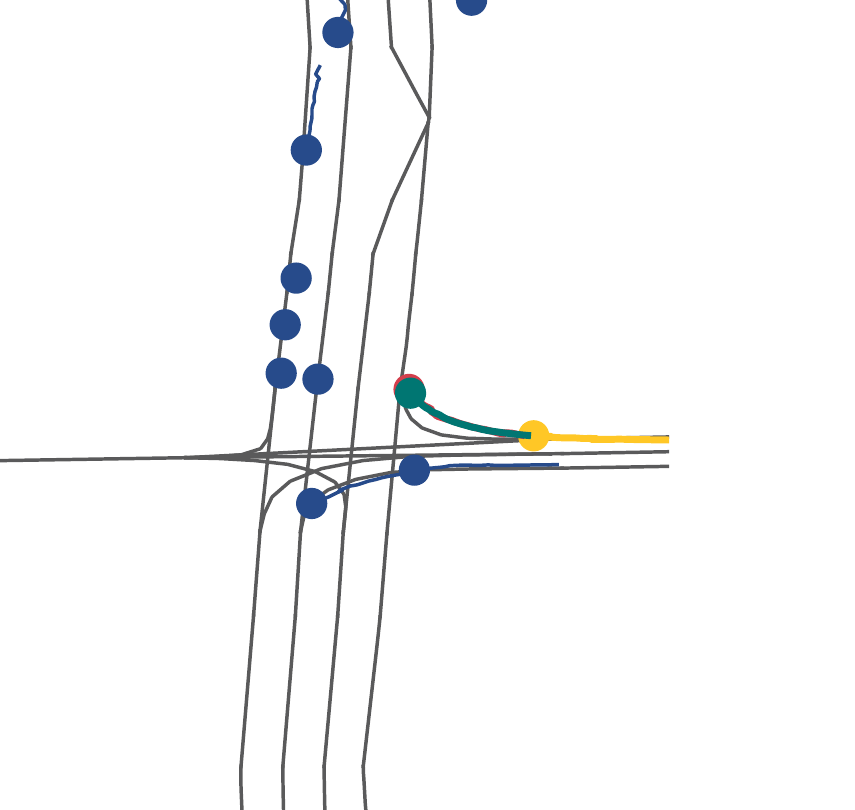}}
\subfloat{\includegraphics[width=0.47\linewidth]{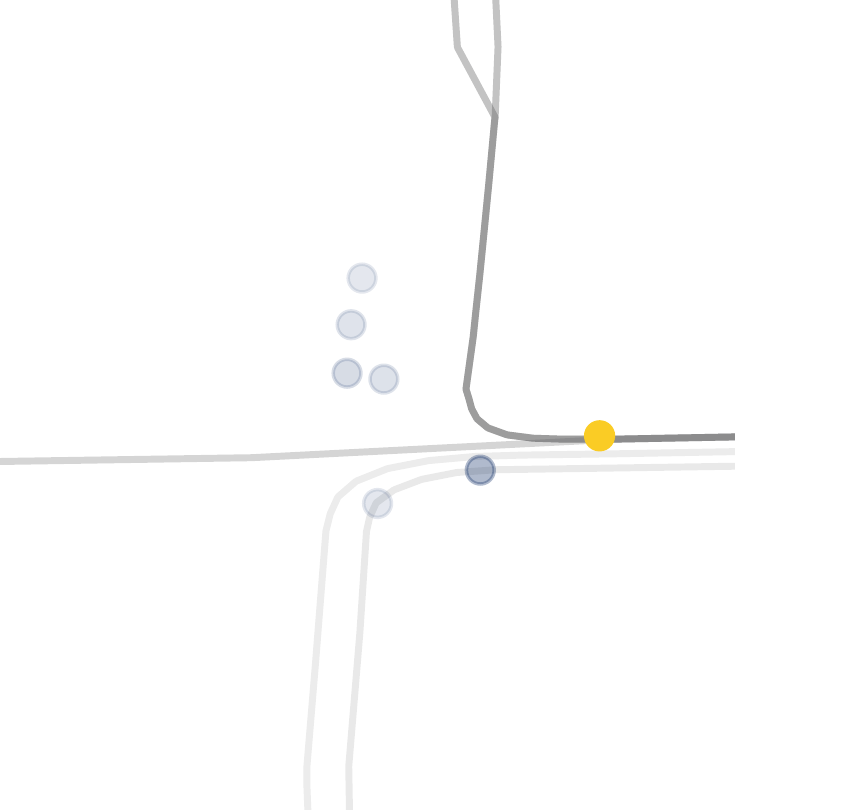}}\\
\vspace{1mm}
\subfloat{\includegraphics[width=0.47\linewidth]{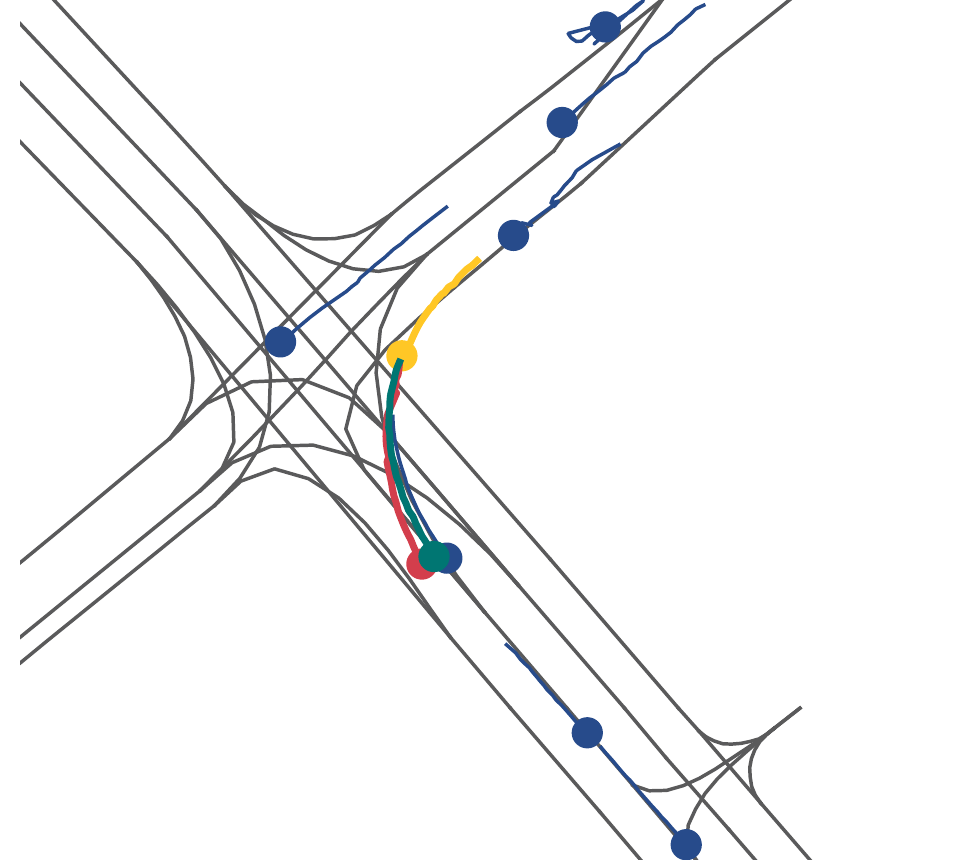}}
\subfloat{\includegraphics[width=0.47\linewidth]{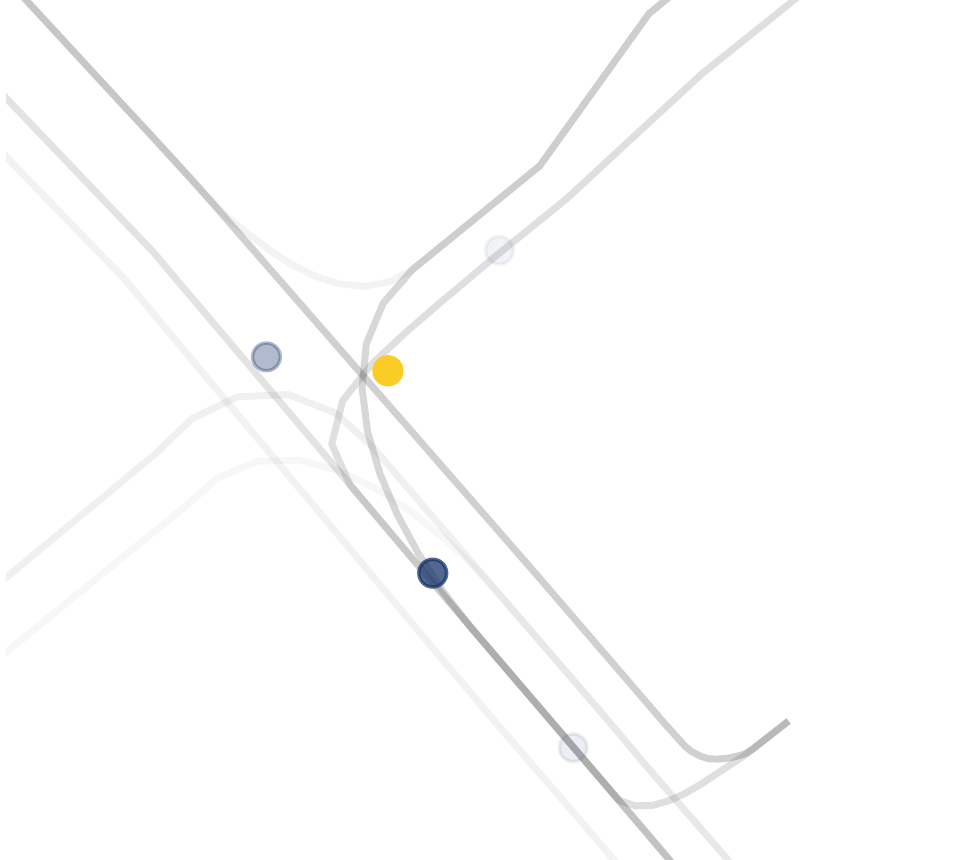}}\\
 \vspace{1mm}
\subfloat{\includegraphics[width=0.49\linewidth]{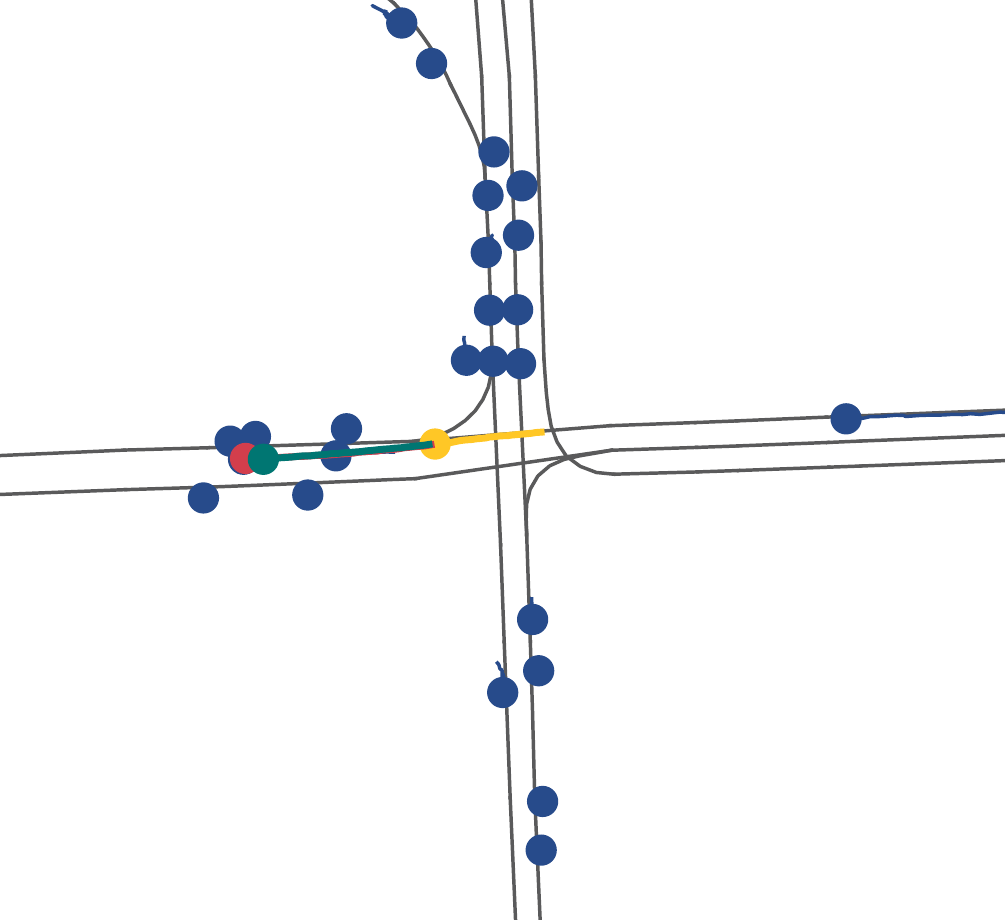}}
\subfloat{\includegraphics[width=0.49\linewidth]{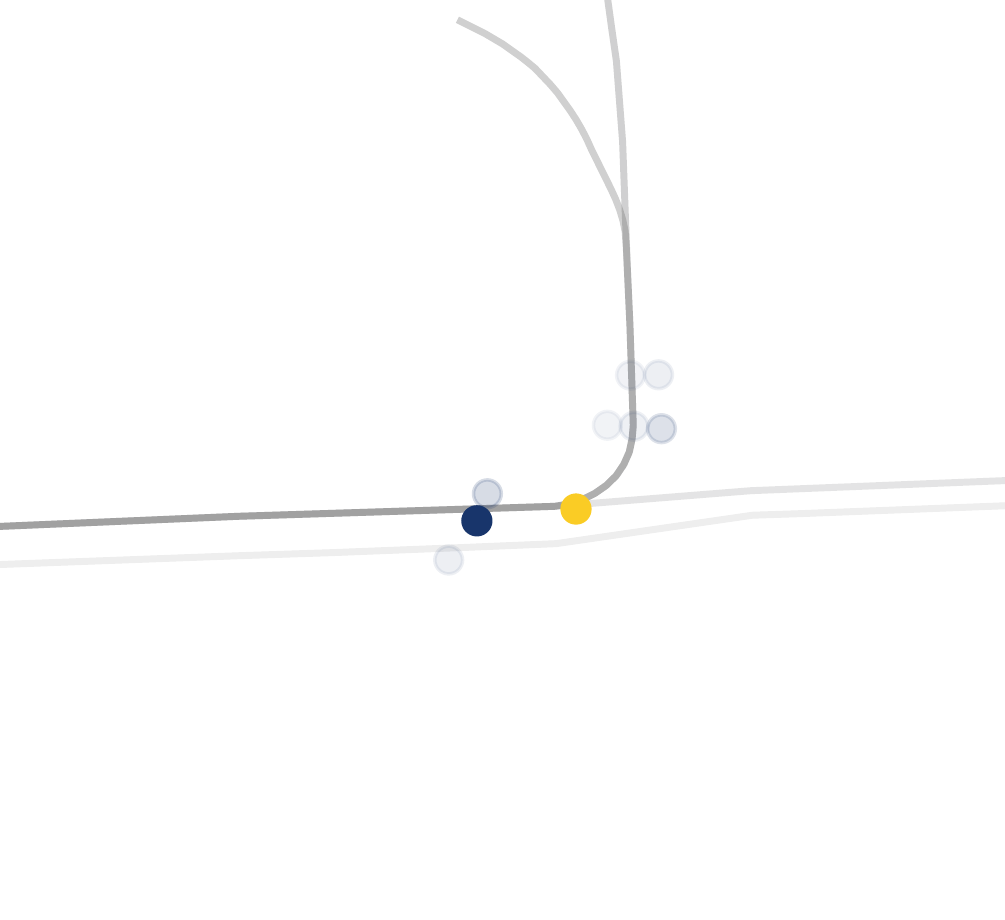}}\\
\vspace{1mm}
\subfloat{\includegraphics[width=0.48\linewidth]{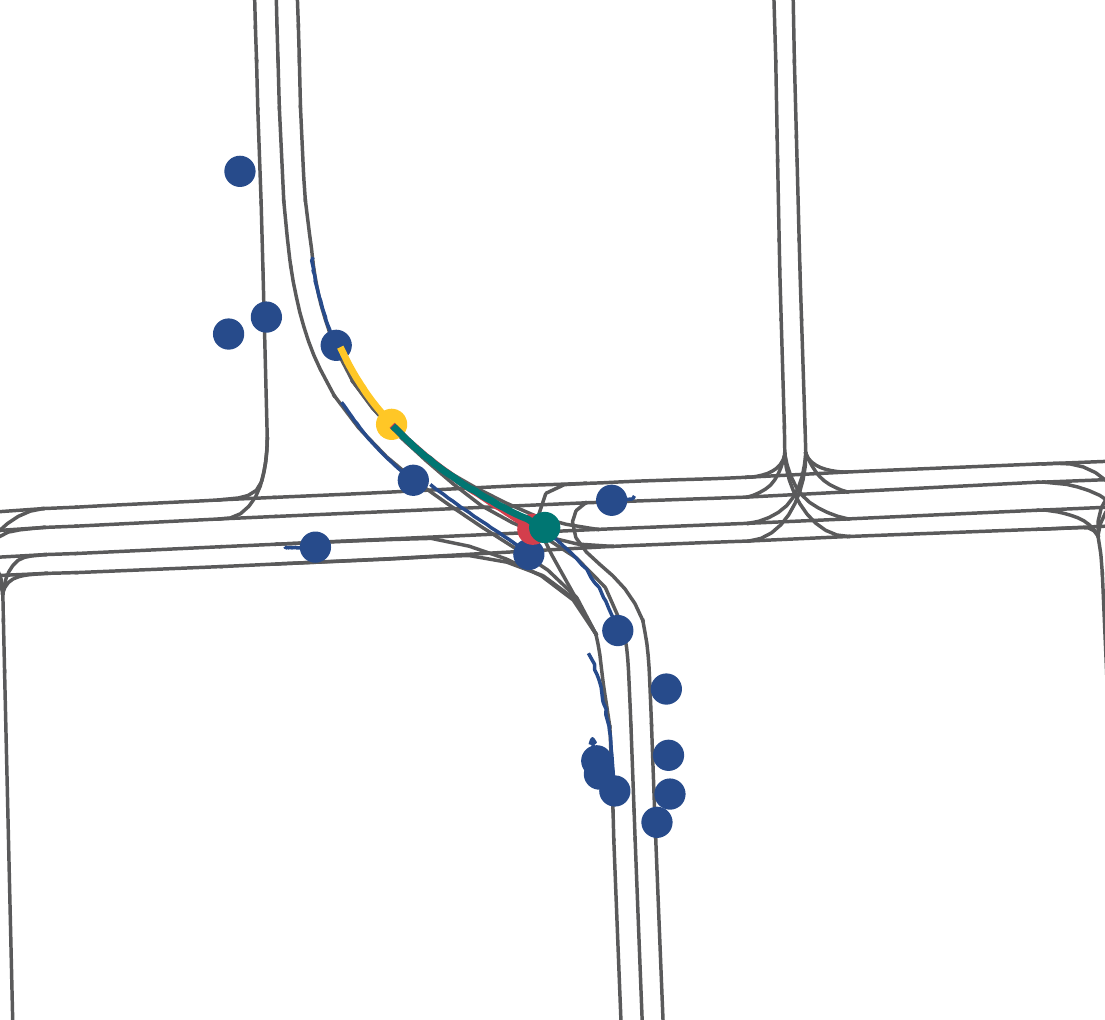}}
\subfloat{\includegraphics[width=0.48\linewidth]{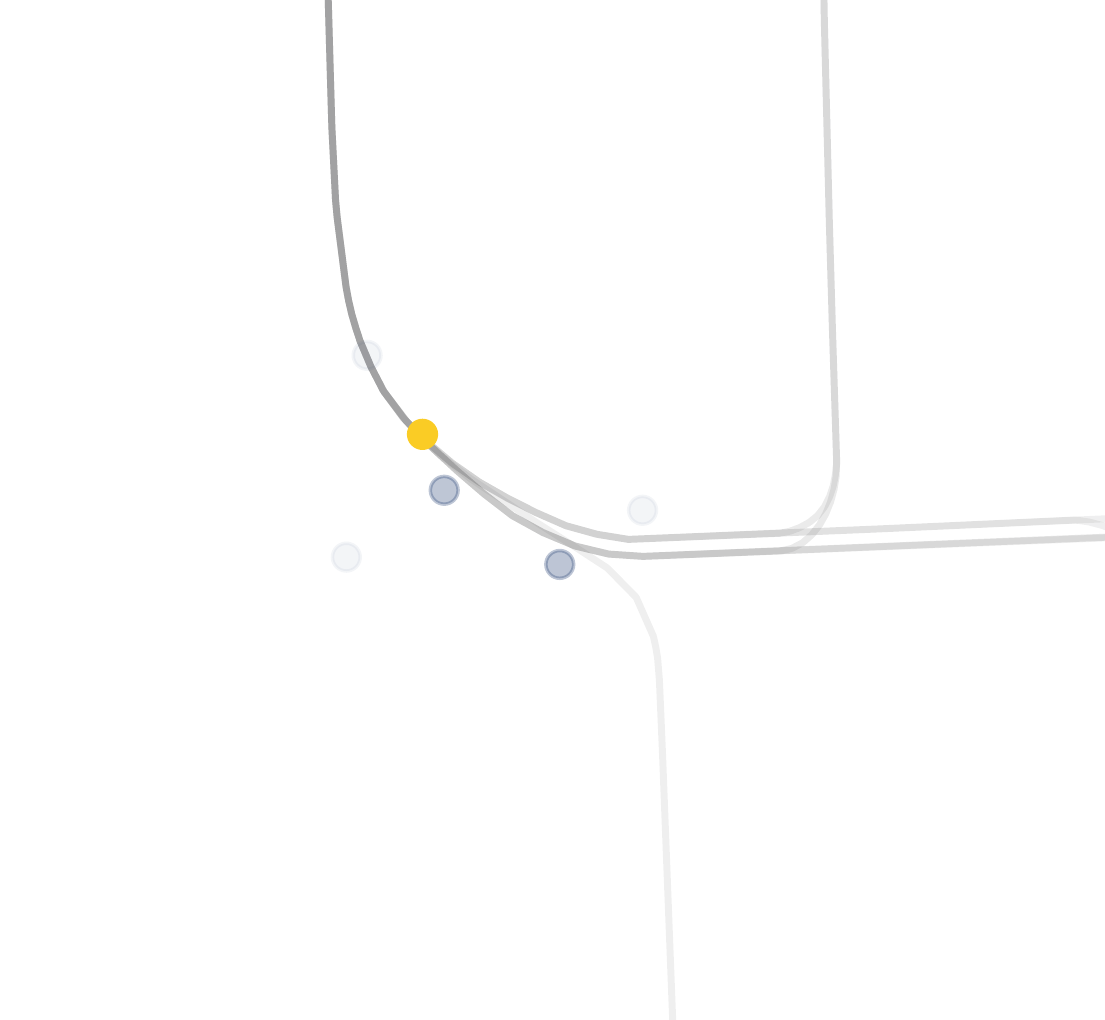}}
\vspace{1mm}
\subfloat{\includegraphics[width=0.48\linewidth]{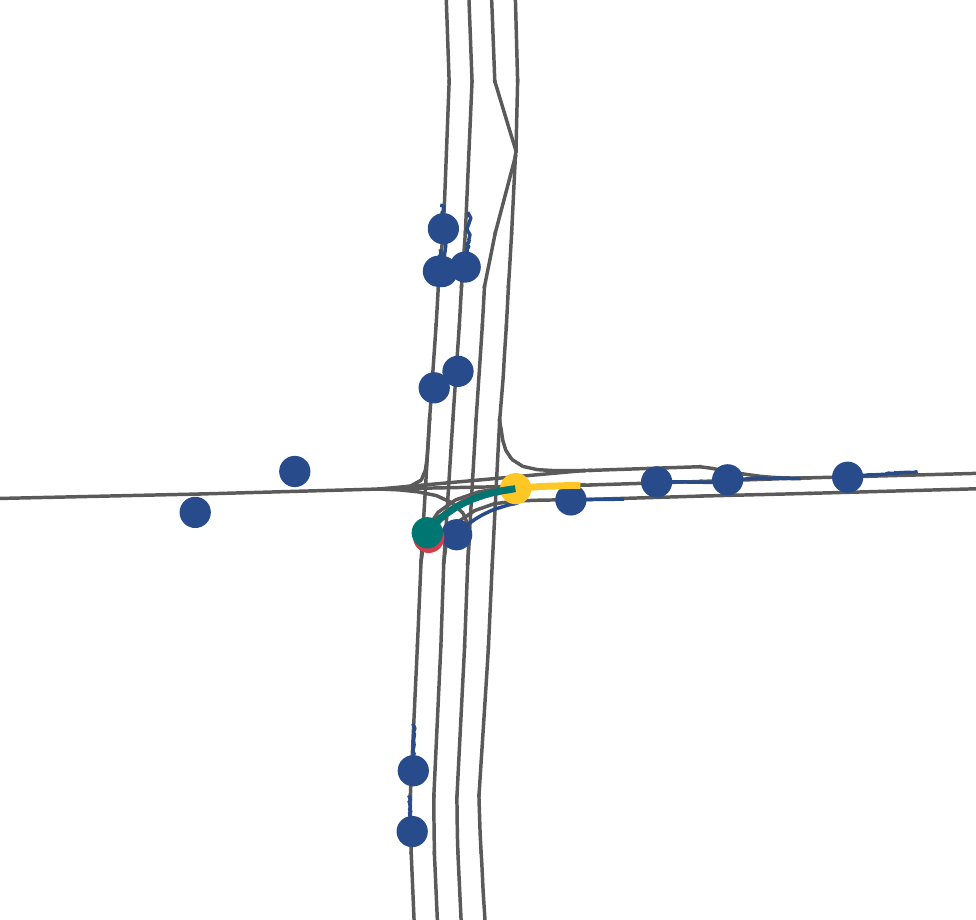}}
\subfloat{\includegraphics[width=0.48\linewidth]{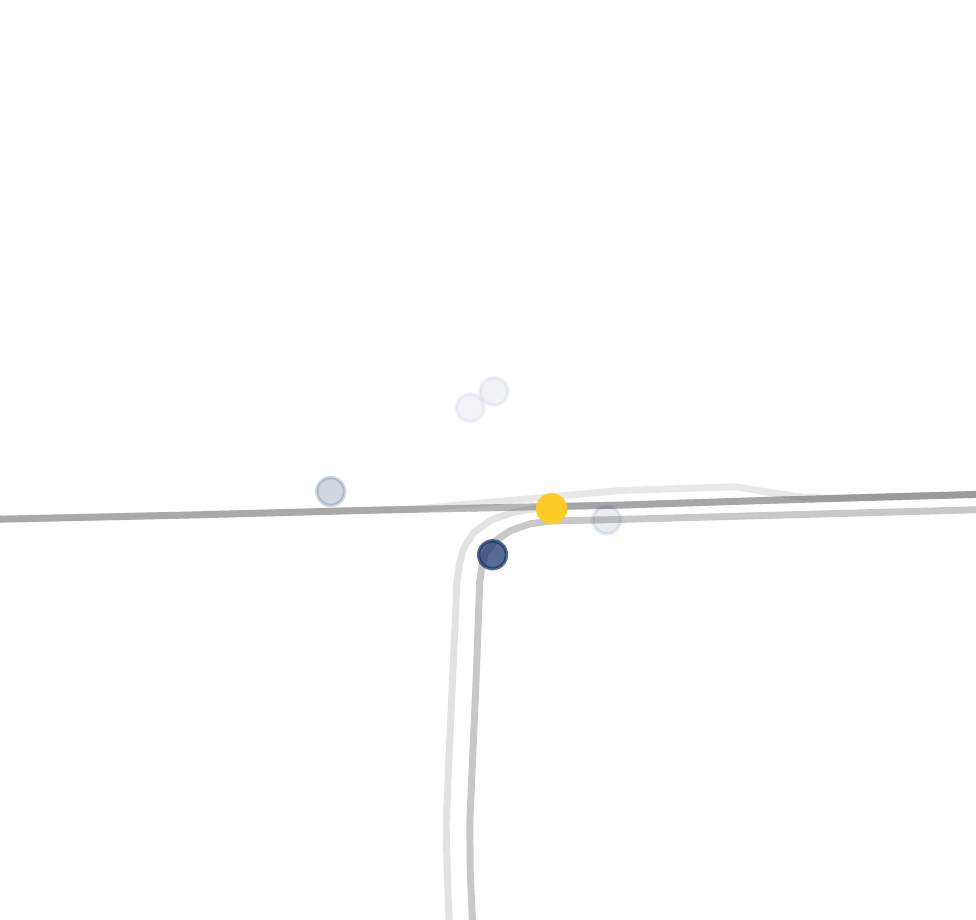}}
\end{center}
\vspace{-15mm}
\hspace{26mm}
\subfloat{\includegraphics[width=0.4\linewidth]{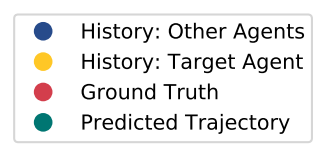}
}
\caption{Qualitative results of SVG-Net. The predictions of the model are on left and the attended regions and agents are on right. Dark blue are other agents along with their observations, yellow shows the history of the main agent, red is the ground truth and green is the prediction. We visualize higher attention values with more opacity. }
\label{fig:qualitative}
\end{figure}

\end{document}